\documentclass{article}%
\usepackage{amsmath}
\usepackage{amsfonts}
\usepackage{amssymb}
\usepackage[margin=1.1in,nohead]{geometry}
\usepackage{setspace,color}
\usepackage{algorithm, algpseudocode}
\usepackage{graphicx}%
\providecommand{\U}[1]{\protect\rule{.1in}{.1in}}
\newcommand{\norm}[1]{\left\lVert#1\right\rVert}
\usepackage{amssymb,amsmath,graphicx,float}
\usepackage{algorithm, algpseudocode}
\usepackage{booktabs}
\usepackage{amsthm}
\usepackage{wrapfig}
\usepackage{setspace}
\usepackage{graphicx}
\usepackage{epsfig}
\usepackage[all]{xy}
\usepackage{verbatim}
\usepackage{float}
\usepackage{mathrsfs}  
\usepackage{bm}
\usepackage{multirow}
\usepackage{color,soul}
\usepackage{amssymb,amsmath,graphicx}
\usepackage{amsthm}

\newtheorem{definition}{Definition}

\newcommand{\tabincell}[2]{\begin{tabular}{@{}#1@{}}#2\end{tabular}}

\usepackage{draftwatermark}
\SetWatermarkText{MANUSCRIPT}
\SetWatermarkScale{2.5}

\begin{document}

\title{Census Signal Temporal Logic Inference for Multi-Agent Group Behavior
	Analysis}

\author{Zhe~Xu
	and Agung~Julius
	\thanks{Zhe~Xu and Agung~Julius are with the Department of Electrical, Computer, and Systems
		Engineering, Rensselaer Polytechnic Institute, Troy, NY 12180, USA e-mail: xuz8,juliua2@rpi.edu.}
}

\date{}
\maketitle

\begin{abstract}
	In this paper, we define a novel census signal temporal logic (CensusSTL)              
	that focuses on the number of agents in different subsets of a group
	that complete a certain task specified by the signal temporal
	logic (STL). \textcolor{black}{CensusSTL consists of an ``\emph{inner logic}'' STL formula and an ``\emph{outer logic}'' STL formula.} We present a new inference algorithm
	to infer CensusSTL formulae \textcolor{black}{from the trajectory data of \textcolor{black}{a group of} agents}. We first identify the ``\emph{inner logic}'' STL formula and then \textcolor{black}{infer the subgroups} based on whether the agents' behaviors satisfy the                               
	``\emph{inner logic}'' formula at each time point. We use two different approaches to
	\textcolor{black}{infer the subgroups} based on similarity and complementarity,
	respectively. The ``\emph{outer logic}''
	CensusSTL formula \textcolor{black}{is} inferred from \textcolor{black}{the census trajectories of different subgroups.} We apply the
	algorithm in analyzing \textcolor{black}{data from a soccer match} by inferring the
	CensusSTL formula for different subgroups of a soccer team.
\end{abstract}

\normalsize{                                                                                                
	\section{Introduction}
	\textcolor{black}{In some multi-agent systems, there are subgroups that perform
		different tasks, such as the defenders, midfielders and forwards 
		in a soccer team} \cite{BLSX2006}. Within each subgroup, the agents can be seen as
	interchangeable members in the sense that as long as there \textcolor{black}{is a
		certain} number of agents in the subgroup performing the task, it
	does not matter who these agents are. In the social network context,                       
	the behavior pattern of different groups of people and the temporal
	influence on them have  been a research focus
	\cite{QYZT2014}, \cite{GHJXH2013}, \cite{ZCDMC2001}.} A recommender system can use \textcolor{black}{this information}
to give better recommendations of the place and time for doing
certain \textcolor{black}{activities} whether it is shopping or checking in at a hotel. In robotics, the Multi-Agent \textcolor{black}{Robot} Systems (MARS)
\cite{SFT1993}, \cite{TKB1999}, \cite{VHKJ2007}, \cite{FVPL2007} are being studied for their co-operative                    
behaviors such as the leader robot \textcolor{black}{tracking} a prescribed
trajectory and the rest of the robots \textcolor{black}{following} the leader while forming
a desired formation pattern \cite{HPA2011}. In all of these
applications, how to express and characterize the properties of the
group behavior has always been a challenge. 

\subsection{Related Works} 

There has been rich literature on formalization of multi-agent group behaviors. In \cite{CanWang}, the authors propose an ontology-based behavior
modeling and checking system to explicitly
represent and verify complex group behavior interactions.
Temporal logic is
a formal approach that has been increasingly used in expressing more
complicated and precise high-level control specifications \cite{KTS2007}, \cite{QJ2006}. There
has been many different temporal logic
frameworks in multi-agent systems to guarantee safe and satisfactory
performance from high level perspectives, such as LTL \cite{FJH2015}, \cite{FDK2012}, CTL \cite{Konur09}, ATL \cite{Alur2002ATL}, etc. The temporal logic formulae are
predefined as a specification for the behaviors of the system \cite{MJD2014}, \cite{WTAC2013}.

Recently,
there is a growing interest in devising algorithms to identify dense-time temporal
logic formulae from system trajectories \cite{zhe2015}. \textcolor{black}{In \cite{Asarin2012}, the authors present a method to synthesize magnitude and timing parameters in a quantitative temporal logic formula so that it fits observed data.} In \cite{KZJA2014}, the authors designed an inference
algorithm that can automatically construct signal temporal logic
formulae directly from data. The obtained signal
temporal logic formulae can be used to classify different behaviors,
predict future behaviors and detect anomaly behaviors \cite{JZB2014}. 

\subsection{Contributions and Advantages} 
In this paper, we define a novel census signal temporal logic (CensusSTL)           
that focuses on the number of agents and the structure of the group
that complete a certain task specified by the signal temporal
logic. The word \textquotedblleft\emph{census }\textquotedblright means \textquotedblleft\emph{the procedure of systematically
	acquiring and recording information about the members of a given
	population}\textquotedblright \cite{weeks1992population}. In the group behavior analysis, we need to generate
knowledge about the behaviors of the members or agents of different
subgroups, and census signal temporal logic provides a formal structure for
generating such knowledge. The census signal temporal logic formula is essentially a signal temporal logic formula (\textquotedblleft\emph{outer logic}\textquotedblright) with the variable in the predicate being the number of      
agents whose behaviors satisfy another signal temporal logic
formula (\textquotedblleft\emph{inner logic}\textquotedblright). For example, the census signal temporal logic formula can express   
specifications such as \textquotedblleft\emph{From 10am to 2pm, at least 3 policemen should be present at the lobby for at least 20 minutes in every hour}\textquotedblright\textcolor{black}{,} where the ``\emph{inner logic}'' formula is the task \textquotedblleft\emph{be present at the lobby for at least 20 minutes in every hour}\textquotedblright and the ``\emph{outer logic}'' formula is \textquotedblleft\emph{from 10am to 2pm, at least 3 policemen should perform the task}\textquotedblright.  

\textcolor{black}{CensusSTL is different from the other Temporal Logic frameworks for multi-agent systems as it does not focus on individual agents or the interaction between different agents, but on the number of agents in different subgroups that complete a certain task. Therefore, it is more useful in applications where only the number of agents or the proportion of agents in different subgroups of a population is of interest while different agents in a subgroup can be seen as interchangeable}.

We present a new inference algorithm that can infer the CensusSTL formula directly from individual agent trajectories. \textcolor{black}{Our inference method for the ``\emph{inner logic}'' formula and the ``\emph{outer logic}'' formula are similar \textcolor{black}{to} \cite{Asarin2012} as we also choose the template of 
	formula first and then search for the parameters. However we formulate the problem as a group behavior analysis problem, so the objective of our approach is not only finding the parameters that fit certain temporal logic formula, but also infer the subgroups and the temporal relationship among the different subgroups.} 

\subsection{Organizations}
This paper is structured as follows. Section II introduces the framework of census signal temporal logic.
Section III shows the algorithm to infer the census temporal
logic formula from data. Section IV implements the algorithm on analysing a            
soccer match as a case study. Finally, some conclusions
are presented in Section V.

\section{Census Signal Temporal Logic}   
\subsection{\textcolor{black}{``\emph{Inner Logic}'' Signal Temporal Logic}}    
In this paper, we find subgroups of a population that act collaboratively for a task. We need to find both the task and the subgroups from the time-stamped trajectories \textcolor{black}{(for definition of time-stamped trajectories, see the beginning of Section II-B)} of different agents. The task can be formulated as an ``\emph{inner logic}'' STL formula.  
Assume there is a group $S$ of $n$ agents and each agent has an observation space $\mathbb{X}$. For example, the group can be a set of points moving in 2D plane, and the observation space can be their 2D positions. Each element of the observation space is described by a set of \textcolor{black}{$w$} variables
that can be written as a vector $x=[x_{1},x_{2},\dots,\textcolor{black}{x_{w}}]^{T}$.  
The domain of $x$ is denoted by $\mathbb{X}=\mathbb{X}_{1}\times\mathbb{X}_{2}\times 
\dots\times\textcolor{black}{\mathbb{X}_{w}}$. The domain $\mathbb{B} = $\{\textcolor{black}{true, false}$\}$ is the
Boolean domain and the time set is $\mathbb{T} = \mathbb{R}$ \textcolor{black}{(note that we allow negative time to add more flexibility of the temporal operator)}. With a
slight abuse of notation, we define  
observation trajectory (or signal or behavior) $x$ describing the observation of
each agent as a function from $\mathbb{T}$ to $\mathbb{X}$.  Therefore, $x_{i}$ refers to both the name of the $i$-th observation                                                                                          
variable and its valuation in $\mathbb{X}$. \textcolor{black}{A finite set $M=\{\mu_{1},\mu
	_{2},\textcolor{black}{\dots\mu_{q}}\}$ is a set of atomic predicates, each mapping \textcolor{black}{$\mathbb{X}$} to $\mathbb{B}$.} The ``\emph{inner logic}'' is signal temporal logic \cite{DAM2010} and the syntax of the ``\emph{inner logic}'' STL proposition can be defined recursively as follows:
\[
\phi:=\top\mid\textcolor{black}{\mu}\mid\lnot\phi\mid\phi_{1}\wedge\phi_{2}\mid\phi_{1}\vee
\phi_{2}\mid\phi_{1}\mathcal{U}_{I}\phi_{2}%
\]
where $\top$ stands for the Boolean constant \textcolor{black}{true}, \textcolor{black}{$\mu$ is an atomic predicate
	in the form of an inequality $f(x(t))> 0$ where $f$ is some real-valued function}, $\lnot$ (negation), $\wedge$(conjunction), $\vee$ (disjunction)
are standard Boolean connectives, $\mathcal{U}$ is a temporal operator
representing \textquotedblleft until\textquotedblright, $I$ is an interval of
the form $I=(i_{1},i_{2}),(i_{1},i_{2}],[i_{1},i_{2})$ or $[i_{1},i_{2}]$ $\textcolor{black}{(i_{1}\le i_{2}, i_{1},i_{2} \in \mathbb{T})}$. \textcolor{black}{In general, a predicate can be an atomic predicate or atomic predicates connected with standard Boolean connectives.}
We   
can also derive two useful temporal operators from                        \textquotedblleft                                      
until\textquotedblright($\mathcal{U}$), which are \textquotedblleft
eventually\textquotedblright$\Diamond\phi=\top\mathcal{U}\phi$ and
\textquotedblleft always\textquotedblright$\Box\phi=\lnot\Diamond\lnot\phi$.                               

We use $(x,t)$ to represent the observation trajectory $x$ at time $t$, then the Boolean semantics of ``\emph{inner logic}'' are defined recursively as follows:
\[%
\begin{split}
\textcolor{black}{(x,t)\models\mu\quad\mbox{iff}\quad}& \textcolor{black}{f(x(t))> 0}\\
(x,t)\models\lnot\phi\quad\mbox{iff}\quad &  (x,t)\nvDash\phi\\
(x,t)\models\phi_{1}\wedge\phi_{2}\quad\mbox{iff}\quad &  (x,t)\models\phi
_{1}\quad and\quad(x,t)\models\phi_{2}\\
(x,t)\models\phi_{1}\vee\phi_{2}\quad\mbox{iff}\quad &  (x,t)\models\phi
_{1}\quad or\quad(x,t)\models\phi_{2}\\
(x,t)\models\phi_{1}\textcolor{black}{\mathcal{U}}_{[a,b)}\phi_{2}\quad\mbox{iff}\quad &  \exists
t^{\prime}\in\lbrack t+a,t+b)\\
&  s.t.(x,t^{\prime})\models\phi_{2},(x,t^{\prime\prime})\models\phi_{1}\\
&  \forall t^{\prime\prime}\in\lbrack t+a,t^{\prime})
\end{split}
\]

The robustness degree of an observation trajectory $x$ with respect to an ``\emph{inner logic}'' formula
$\phi$ at time $t$ is given as $r(x,\phi,t)$, where $r$ can be calculated
recursively via the quantitative semantics \cite{DAM2010}:%
\[%
\begin{split}
r(x,\mu,t)  &  =f(x(t)),\\
r(x,\lnot\phi,t)  &  =-(r(x,\phi,t)),\\
r(x,\phi_{1}\wedge\phi_{2},t)  &  =\min(r(x,\phi_{1},t),r(x,\phi_{2},t)),\\
r(x,\phi_{1}\vee\phi_{2},t)  &  =\max(r(x,\phi_{1},t),r(x,\phi_{2},t)),\\
r(x,\Box_{\lbrack\tau_{1},\tau_{2})}\phi,t)  &  =\min\limits_{t+\tau_{1}\leq
	t^{\prime}<t+\tau_{2}}r(x,\phi,t^{\prime}),\\ 
r(x,\Diamond_{\lbrack\tau_{1},\tau_{2})}\phi,t)  &  =\max\limits_{t+\tau
	_{1}\leq t^{\prime}<t+\tau_{2}}r(x,\phi,t^{\prime})
\end{split}
\]
\textcolor{black}{\[%
	\begin{split}
	r(x,\phi_{1}\mathcal{U}_{[a,b)}\phi_{2},t)  &  =\sup\limits_{t+a\leq t^{\prime}<t+b}(\min(r(s,\phi_2,t^{\prime}),\\&\sup\limits_{t\leq t^{\prime\prime}<t^{\prime}}r(s,\phi_1,t^{\prime\prime})))\end{split}
	\]}

\subsection{\textcolor{black}{Signal Temporal Logic Applied to Data}}  

We make two deviations to STL when applying an ``\emph{inner logic}'' formula
$\phi$ to data: \\
1) As the observation trajectory is usually of finite length, \textcolor{black}{and also considering that there may be negative time in the temporal operator of the ``\emph{inner logic}'' formula $\phi$}, \textcolor{black}{the satisfaction of the ``\emph{inner logic}'' formula $\phi$ may not be well-defined} at every time point of the observation trajectory (for example for the formula $\phi_1=\Diamond_{[0,10)}(x>5)$ and \textcolor{black}{$\phi_2=\Diamond_{[-10,0)}(x>5)$}, if the observation trajectory is defined on the time domain of $[0,200]$, then $\phi_1$ can only be evaluated on the time domain of [0, 190] and \textcolor{black}{$\phi_2$ can only be evaluated on the time domain of [10, 200]}). Assume that the time domain of the observation trajectory $x$ is $\mathbb{T}_o \subset \mathbb{T}$, \textcolor{black}{with a slight abuse of notation, we define time-stamped trajectory $x$ of finite length as a function from $\mathbb{T}_o$ to $\mathbb{X}$.}

The time domain of the ``\emph{inner logic}'' formula $\phi$ with respect to $x$ is defined recursively as follows:                                                                                          
\[%
\begin{split}
\textcolor{black}{D(\mu, x)}& \textcolor{black}{=\mathbb{T}_o}\\
\textcolor{black}{D(\lnot\phi, x)} & \textcolor{black}{=D(\phi, x)}\\
\textcolor{black}{D(\phi_{1}\wedge\phi_{2}, x)} & \textcolor{black}{=D(\phi_{1}, x)\cap D(\phi_{2}, x)}\\
\textcolor{black}{D(\phi_{1}\mathcal{U}_{[a,b)}\phi_{2}, x)} & \textcolor{black}{=\{t \mid [t+a, t+b) \subset (D(\phi_{1}, x)\cap D(\phi_{2}, x))\}}
\end{split}
\]

For example, for ``\emph{inner logic}'' formula $\phi=\Diamond_{[0,10)}(x>5)\wedge \Diamond_{[20,40)}(\Box_{[20,60)}(x>20))$\textcolor{black}{, if} the observation trajectory is defined on $[0,200]$, then $D(\phi)$=[0-0, 200-10]$\cap$[0-20-20, 200-40-60]=[0, 100]. 

\textcolor{black}{2) The observation data \textcolor{black}{are} usually discrete, so the time domain of the observation trajectory  $\mathbb{T}_o$ is a set of discrete time points. In this case, the interval $I$ in 
	the form $I=(i_{1},i_{2}),(i_{1},i_{2}],[i_{1},i_{2})$ or $[i_{1},i_{2}]$ actually means the time points in $\mathbb{T}_o$ that belongs to $I$. For example, $[i_{1},i_{2})$ is interpreted as $\{t \in \mathbb{T}_o \mid t \in [i_{1},i_{2})\}$.}

\begin{figure}[ht]
	\centering
	\includegraphics[width=9cm]{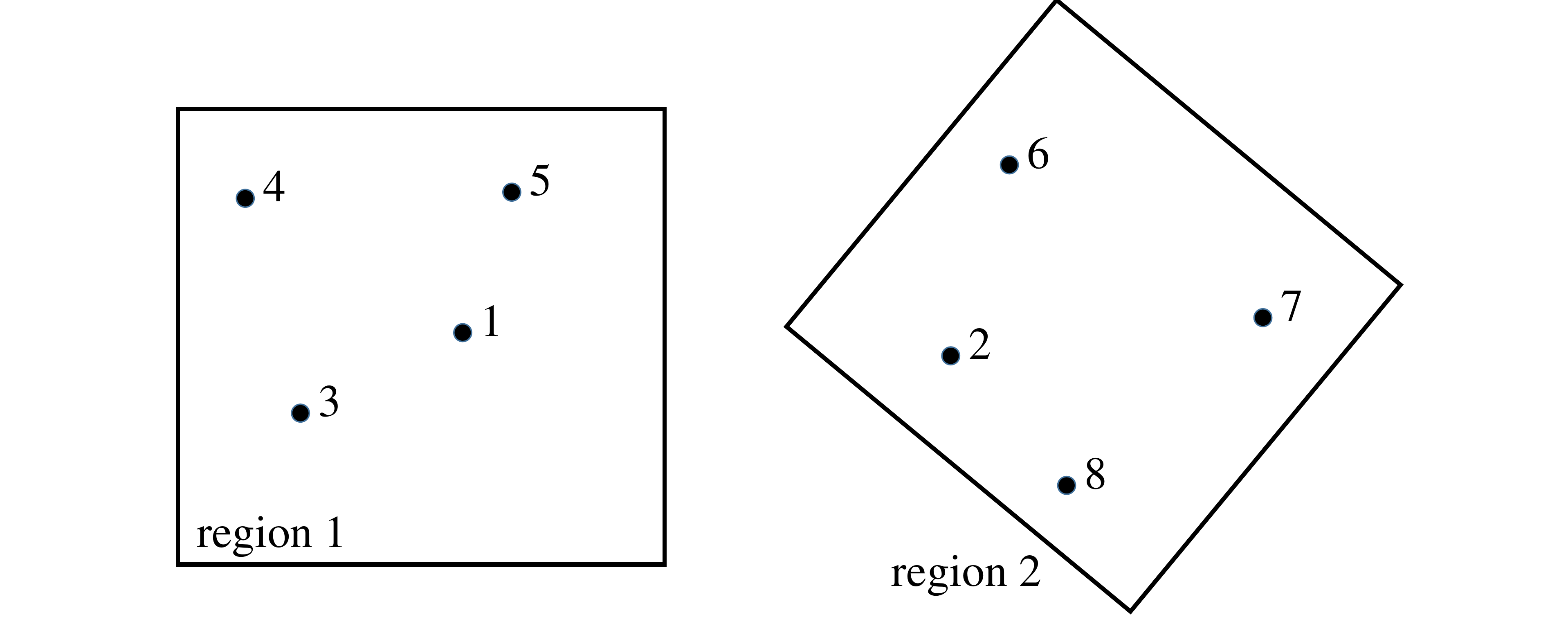}
	\caption{8 agents and 2 regions in the furniture moving example.}
	\label{Room}%
\end{figure}
\textcolor{black}{Consider
	the example shown in Fig.~\ref{Room} where there are two predicates \textcolor{black}{$p_{region1}$ and $p_{region2}$ corresponding to region 1 and region 2 (for the representation of predicates, see Eq. (4) in Section III)} and 8 different agents (people) who are moving furnitures from Region 1 to Region 2. The people need to move back and forth frequently between Region 1 and Region 2.
	One STL formula that can characterize the \textcolor{black}{people moving pattern} is}
\begin{align}
\textcolor{black}{\phi_t=}&\textcolor{black}{\Box_{[0,\tau_1)} \textcolor{black}{p_{region1}} \wedge \Diamond_{[\tau_2,\tau_3)} \textcolor{black}{(p_{region2}} \wedge } \textcolor{black}{\Diamond_{[\tau_4,\tau_5)} \textcolor{black}{p_{region1})}}   
\end{align}
which reads \textquotedblleft\emph{the person is in region 1 for $\tau_1$ time units and arrives in region 2 sometime between $\tau_2$ and $\tau_3$ time units, then sometime between $\tau_4$ and $\tau_5$ time units \textcolor{black}{later} the person comes back to region 1}\textquotedblright. The temporal parameters satisfy \textcolor{black}{$\tau_5 \ge \tau_4$, $\tau_3\ge \tau_2$, $\tau_1\ge 0$}.

\textcolor{black}{We specify that the ``\emph{inner logic}'' formula $\phi=\Diamond_{[-\tau_5\textcolor{black}{-\tau_3},0]}\phi_t$ while the temporal operator $\Diamond_{[-\tau_5\textcolor{black}{-\tau_3},0]}$ is to make $\phi$ true at every time point during the execution of the task. Without this temporal operator, $\phi$ is only true at the beginning of the task}.

\subsection{\textcolor{black}{``\emph{Outer logic}'' Census Signal Temporal Logic}}  
Based on the ``\emph{inner logic}'', we can define the ``\emph{outer logic}'' census signal temporal logic (CensusSTL).
The observation element of the ``\emph{outer logic}'' is the number of agents that satisfy the ``\emph{inner logic}'' formula, which can be described by non-negative integers that belong to the domain $\mathbb{N}$. A census trajectory $n(\phi,S)$ describes the number of agents in the group $S$ whose behaviors satisfy the ``\emph{inner logic}'' formula $\phi$ \textcolor{black}{over time}. \\        

It should be noted that the time domain of the census trajectories is the same as the time domain of the ``\emph{inner logic}'' formula $\phi$ with respect to observation trajectory $x$ if the observation trajectory is of finite length, so \textcolor{black}{$n(\phi,S)$ is a mapping from $D(\phi, x)$ to $\mathbb{N}$}. \textcolor{black}{As there may be different subgroups $S_i(i=1,2,\dots)$ in group $S$, we have the following definition.}

\begin{definition}
	We define $n(\phi,S_i,t)$ as the number of agents
	in the subgroup $S_i(i=1,2,\dots)$ whose behaviors satisfy the ``\emph{inner logic}'' formula $\phi$ at time $t$, \textcolor{black}{or} in other words, the number of agents whose behavior (observation trajectory) has positive robustness degree with respect to $\phi$ at time $t$.
\end{definition}

With that notation, the atomic \textcolor{black}{predicate} of the ``\emph{outer logic}'' CensusSTL can be defined as follows,
\begin{equation}
\textcolor{black}{\mu_n :=n(\phi,S_i)> c\mid-n(\phi,S_i)> -c }   
\label{gamma}                 
\end{equation} 
where $c$ is a non-negative integer.

\textcolor{black}{ In the furniture moving example, assume that there are two subgroups of people who are \textcolor{black}{moving the furnitures}, $\{1,2,3,4\}$ and $\{5,6,7,8\}$. The atomic predicate of the ``\emph{outer logic}'' can express properties such as \textquotedblleft\emph{the number of people in the subgroup $\{1,2,3,4\}$ who are moving the furnitures is less than 2}\textquotedblright, or \textquotedblleft\emph{the number of people in the subgroup $\{5,6,7,8\}$ who are moving the furnitures is more than 3}\textquotedblright. }      

The syntax of the ``\emph{outer logic}'' CensusSTL \textcolor{black}{proposition} can be defined recursively as follows:
\begin{align}  
\gamma := \top \mid \mu_n \mid \neg \gamma \mid \gamma_1\wedge\gamma_2 \mid \gamma_1\vee\gamma_2 \mid \gamma_1 \mathcal{U}_I\gamma_2
\label{eq:gamma}
\end{align}

As the ``\emph{outer logic}'' CensusSTL is also STL, so the semantics of STL also \textcolor{black}{applies} to the ``\emph{outer logic}'' CensusSTL. The robustness degree of \textcolor{black}{a} census trajectory $n$ with respect to \textcolor{black}{a} CensusSTL formula $\gamma$ at
time $t$ is \textcolor{black}{denoted} as $r(n,\gamma, t)$, where $r$ can be
calculated recursively in the same way as $r(x,\phi,t)$ is calculated.

\section{Census Signal Temporal Logic Inference} 

\textcolor{black}{In this section, we seek} to infer the CensusSTL formula \textcolor{black}{describing the behaviors of a group of agents} from \textcolor{black}{the collection of the individual agent observation trajectories in a training data set and then test the validity of the inferred CensusSTL formula in a separate validation data set}. We choose to represent the \textcolor{black}{predicates} as polyhedral sets as \textcolor{black}{they} are more general than rectangular sets and computationally easier to handle than other more complex sets (ellipsoidal sets, non-convex sets, etc.). So each predicate in the ``\emph{inner logic}'' formula is represented in the following form: 

\begin{equation}
p:=\left(\bigwedge_{k=1}^{m}a_{k}^{T}x>b_{k}\right)
,a_{k}\in\mathbb{R}^{n},b_{k}\in\mathbb{R,}%
\end{equation}
where vector ${a}_k$ and number $b_k$ denote the                    
parameters that define the predicate, $m$ is the number of atomic predicates in the           
predicate. 


\textcolor{black}{According to the quantitative semantics of STL,} the robustness of each predicate $p$ can be expressed as \textcolor{black}{the minimum of robustness of each atomic predicate}:  

\begin{equation}
\textcolor{black}{r(x,p,t)=\min\limits_{1\le k \le m} (a_{k}^{T}x-b_{k}),~a_{k}\in\mathbb{R}^{n},~b_{k}\in\mathbb{R.}}%
\end{equation}        

\subsection{Task Description}

\textcolor{black}{In this paper, we infer the ``\emph{inner logic}'' STL formula in the form of $\phi=\Diamond_{[-\norm{\phi_t},0]}\phi_t$ where $\phi_t$ is the formula that describes the task with all the temporal parameters of $\phi_t$ chosen in $\mathbb{R}_{\geqslant 0}$ and $\norm{\phi_t}$ is the necessary length associated with formula $\phi_t$ as defined below:}
\[%
\begin{split} 
\textcolor{black}{\norm{\phi_t} := \min\{T\mid \textrm{if}~\mathbb{T}_o=[0, T], D(\phi_t, x)\neq \emptyset\}}
\end{split}
\] 
\textcolor{black}{Take STL formula $\phi_t=\Diamond_{[0,10)}(x>5)\wedge \Diamond_{[20,40)}(\Box_{[20,60)}(x>20))$ for example, the necessary length       $\norm{\phi_t}=40+60=100$， and $\phi=\Diamond_{[-100,0]}\phi_t$ is true at every time point during the execution of the task.
	We consider 4 templates of temporal logic formula $\phi_t$ corresponding to 4 common tasks:}
\subsubsection{Sequential Task}
\begin{align}                                                   
\begin{split}
\textcolor{black}{\phi_t=}&\textcolor{black}{\Box_{[0,\tau_1)} \phi_{t1} \wedge   \Diamond_{[\tau_{21},\tau_{22})}\Box_{[0,\tau_{23})} \phi_{t2}\wedge\dots \wedge}\\&\textcolor{black}{\Diamond_{[\tau_{z1},\tau_{z2})}\Box_{[0,\tau_{z3})} \phi_{tz} }  
\end{split}
\end{align}
\textcolor{black}{where $\phi_{t1}, \phi_{t2}, \dots, \phi_{tz}$ are subtasks that can be predicates or STL formulae as $\phi_{t}$ and the temporal parameters satisfy $\tau_{21}\ge\tau_{1}$, $\tau_{i2}+\tau_{i3}\le\tau_{j1}(2\le i <j\le z)$.
	For any term $\Diamond_{[\tau_{i1},\tau_{i2})}\Box_{[0,\tau_{i3})} \phi_{ti}$ $(i=2,3,\dots,z)$, if $\tau_{i1}=\tau_{i2}$, then this term shrinks to $\Box_{[\tau_{i1},\tau_{i1}+\tau_{i3})} \phi_{ti}$; if $\tau_{i3}=0$, then this term shrinks to $\Diamond_{[\tau_{i1},\tau_{i2})}\phi_{ti}$. The sequential task is a series of subtasks that are performed in a sequential order.}

\subsubsection{Concurrent Task}
\begin{align}
\begin{split}
\textcolor{black}{\phi_t=}&\textcolor{black}{\Box_{[0,\tau_1)}(\phi_{t1} \vee \phi_{t2} \dots \vee\phi_{tz})  }  
\end{split}
\end{align}  
This concurrent task means \textquotedblleft\emph{during the next $\tau_1$ time units, the agent \textcolor{black}{performs at least one of} the subtasks $\phi_{ti}$}\textquotedblright.
\subsubsection{Persistent Task}
\begin{align}
\begin{split}
\textcolor{black}{\phi_t=}&\textcolor{black}{\Box_{[0,\tau_1)}\Diamond_{[0,\tau_2)} \phi_{t1}}  
\end{split}
\end{align}  
\textcolor{black}{ This persistent task means \textquotedblleft\emph{during the next $\tau_1$ time units, $\phi_{t1}$ is performed at least once in every $\tau_2$ time units}\textquotedblright.}
\subsubsection{Causal Task}
\begin{align}
\begin{split}
\textcolor{black}{\phi_t=}&\textcolor{black}{\Box_{[0,\tau_1)}(\phi_{t1} \Rightarrow \phi_{t2})}  
\end{split}
\end{align}  
\textcolor{black}{where $\phi_{t1}$
	is the cause formula and $\phi_{t2}$ is the effect formula. This causal task means \textquotedblleft\emph{during the next $\tau_1$ time units, whenever the subtask $\phi_{t1}$ is performed, the agent will perform subtask $\phi_{t2}$ }\textquotedblright.}

\textcolor{black}{In all of these task templates, we set \textcolor{black}{an} upper limit to the necessary length associated with formula $\phi_t$ as we only consider tasks that is finished within certain time. For example, if it generally takes no more than 10 \textcolor{black}{time units} to move the \textcolor{black}{furniture} from Region 1 to Region 2, then we set $\norm{\phi_t}\le 10$.}

\textcolor{black}{In the following, we introduce the specific steps to infer the CensusSTL formula from data. Note that our procedure cannot produce a formula that does not conform with the predetermined templates. Our aim is to find the CensusSTL formula that best fits (according to some measure of fitness) a given finite set of observation trajectories. Generally, we are given a training data set of $z$ different observation trajectories for each agent, where the time domains of the $z$ observation trajectories are not necessarily the same}.

\subsection{\textquotedblleft{Inner Logic}\textquotedblright STL Formula Inference}

\textcolor{black}{In this section, we discuss the three requirements the ``\emph{inner logic}'' formula needs to meet and then formulate the optimization problem for the ``\emph{inner logic}'' formula inference.}

\subsubsection{Consistency}
We heuristically postulate that if the number of agents whose behaviors satisfy the ``\emph{inner logic}'' formula is changing drastically through time, then \textcolor{black}{the formula} \textcolor{black}{cannot} reflect a task that a group of agents are \textcolor{black}{performing consistently}.\\

\begin{definition}
	We define $v\textcolor{black}{_q}(\phi,S)$ as the \textcolor{black}{temporal} variation
	of the number of agents in the set $S$ whose \textcolor{black}{$q$-th observation trajectories} satisfy the
	``\emph{inner logic}'' formula $\phi$, which can be described as follows:                                             
	\begin{align} 
	\begin{split}
	v\textcolor{black}{_q}(\phi,S) = &\frac{1}{l_{\phi,q}-1}
	\sum_{j=1}^{l_{\phi,q}-1} |n\textcolor{black}{_q}(\phi,S,j+1)-n\textcolor{black}{_q}(\phi,S,j)|\\ 
	\end{split}
	\label{var}
	\end{align}
	where $n\textcolor{black}{_q}(\phi,S,j)$ is the number of agents in the set $S$ whose
	\textcolor{black}{$q$-th observation trajectories} satisfy the STL formula $\phi$ at the $j$-th time point, \textcolor{black}{$l_{\phi,q}$ is the number of time points in the time domain \textcolor{black}{of the $q$-th} census trajectory.}  	
\end{definition}   

\subsubsection{Frequency}
\textcolor{black}{We postulate that if the number of time points at which the behavior of any agent satisfies the
	``\emph{inner logic}'' formula $\phi$ is small, then the formula \textcolor{black}{cannot} reflect a task that is performed frequently}. 

\begin{definition}
	We define $m(\phi,k)$ as the total number of time points at which the
	``\emph{inner logic}'' formula $\phi$ is true for agent $k$ \textcolor{black}{in the training data set} \textcolor{black}{(the time points in different observation trajectories are counted separately)}.
\end{definition}                                         
\subsubsection{Specificity} 
\textcolor{black}{Sometimes a consistent and frequent task can be overly general or meaningless. For example, the proposition \textquotedblleft\emph{the agent is always in the entire space $\mathbb{X}$}\textquotedblright is always true but does not contain any useful information. To make the task more specific and meaningful}, we incorporate some a priori knowledge about the system. \textcolor{black}{The other purpose of incorporating a priori knowledge is to make the task more tailored to the user preferences. For example, if the user is particularly interested in the behavior in a certain region, then this region can be specified as an a priori predicate.} Suppose that we are given a priori predicates
$X_i(i=1,2,\dots,n_p)$, we 
make the obtained predicates $p_i$ \textcolor{black}{as similar as possible}                                              
to the a priori predicates $X_i$. The Hausdorff distance is an
important tool to measure the similarity between two sets of points                                     
\cite{MJ1983}. It is defined as the largest          
distance from any point in one of the sets, to the closest point in
the other set. \textcolor{black}{Suppose that the set of states that satisfy the predicate $p$ is 
	$\mathcal{O}(p)\subset\mathbb{X}$.}  Then
the Hausdorff distance $d_{\mathrm H}(\textcolor{black}{\mathcal{O}(X_i),\mathcal{O}(p_i)})$ is expressed as follows \\
\begin{align}
\begin{split}
d_{\mathrm H}(\textcolor{black}{\mathcal{O}(X_i),\mathcal{O}(p_i)}) =& \max\{\,\sup \limits_{x \in \textcolor{black}{\mathcal{O}(X_i)}} \inf \limits_{y
	\in \textcolor{black}{\mathcal{O}(p_i)}} d(x,y),\\& \sup \limits_{y \in \textcolor{black}{\mathcal{O}(p_i)}} \inf \limits_{x \in \textcolor{black}{\mathcal{O}(X_i)}}
d(x,y)\,\}
\end{split}
\end{align} \\ 

The expression $\sup \limits_{x \in \textcolor{black}{\mathcal{O}(X_i)}} \inf \limits_{y
	\in \textcolor{black}{\mathcal{O}(p_i)}} d(x,y)$ when
both \textcolor{black}{$\mathcal{O}(X_i)$} and \textcolor{black}{$\mathcal{O}(p_i)$} are convex polyhedra can be \textcolor{black}{evaluated} as follows: 

\noindent\textbf{Step 1:} Calculate all vertices of the polyhedron \textcolor{black}{$\mathcal{O}(X_i)$}. Denote
them as $\psi_{1},\psi_{2},\cdots,\textcolor{black}{\psi_{N_{\mathcal{O}(X_i)}}}$.

\noindent\textbf{Step 2:} Calculate the distance from $\psi_{j}$ to $\mathcal{O}(p_i)$ for
each $j\in\{1,\cdots,\textcolor{black}{N_{\mathcal{O}(X_i)}}\}$. This is a convex quadratic optimization problem.

\noindent\textbf{Step 3:} Find the maximum of the distances calculated in Step 2.\\

We denote all parameters
that define the ``\emph{inner logic}'' STL formula $\phi$ as $\alpha$. \textcolor{black}{Take the case of $\phi=\Box_{\lbrack\tau_{1},\tau_{2})}(\bigwedge\limits_{k=1}^{m}a_{k}^{T}x>b_{k})$ for example. As $x+2y>4$ and $2x+4y>8$ are essentially the same,
	we constraint $ \lVert {a}_k \rVert_2$ to be $1$.                              
	One simple way to \textcolor{black}{remove this constraint is to
		represent $a_{k}$ using trigonometric parameters $\theta_{k,1}%
		,\theta_{k,2},\dots$}. \textcolor{black}{Then} $a_{k}^{T}x$ can be represented as $\cos
	(\theta_{k,1})x_{1}+\sin(\theta_{k,1})\cos(\theta_{k,2})x_{2}+\sin(\theta_{k,1})\sin(\theta_{k,2})x_{3}+\dots $ \textcolor{black}{by} utilizing the fact that $\sin^2(\theta_{k,j})+\cos^2(\theta_{k,j})=1.$
	For the formula $\phi(\alpha)$ \textcolor{black}{above}, $\tau_{1},\tau_{2}, \theta_{k,j}, b_{k}$} \textcolor{black}{are} the elements of $\alpha$. The lower bound and upper bound of the angles are set to be $[-\pi,\pi]$.\\

\textcolor{black}{To summarize the three requirements, the inference of the ``\emph{inner logic}'' formula $\phi=\Diamond_{[-\norm{\phi_t},0]}\phi_t$ where $\phi_t$ \textcolor{black}{conforms to 1 of} the 4 task templates is a constrained multi-objective problem, i.e.\\  
	Objectives:\\            
	min $\textcolor{black}{\displaystyle\sum_{q=1}^{z}}v\textcolor{black}{_q}(\phi(\alpha),S)$  (consistency)  \\                   
	max $\displaystyle\sum_{k=1}^{n} m(\phi(\alpha),k)$  (frequency)                                 \\
	min  $\displaystyle\sum_{i=1}^{n_p} d_{\mathrm H}(\mathcal{O}(X_i),\mathcal{O}(p_i(\alpha)))$ (specificity)                                 \\
	Subject to:\\
	$\norm{\phi_t(\alpha)}\le \tau_{limit}$\\
	where $\alpha$ is the optimization variable, $\tau_{limit}$ is the upper limit of the necessary length associated with formula $\phi_t(\alpha)$.}

We use Particle Swarm Optimization \textcolor{black}{\cite{FJZ2013}} to optimize $\alpha$ (including the spatial parameters $\theta, b$ and the temporal parameters $\tau$) of each possible ``\emph{inner logic}'' formula. \textcolor{black}{In each iteration,
	the parameters are updated as a swarm of particles that move in the parameter space to find the global minimum (in this paper, we use 200 particles for each iteration)}.
The formula with the smallest value of the cost function can be generated and selected. The cost function is as follows:
\begin{align}
\begin{split}
J_{stl}(\phi,\alpha) = & 
\textcolor{black}{\sum_{q=1}^{z}}v\textcolor{black}{_q}(\phi(\alpha),S) -\lambda_1\displaystyle\sum_{k=1}^{n} m(\phi(\alpha),k)
+\\&\lambda_2 \displaystyle\sum_{i=1}^{n_p} d_{\mathrm H}(\textcolor{black}{\mathcal{O}(X_i),\mathcal{O}(p_i(\alpha)})
\end{split}   
\label{innercost}
\end{align}
where $\lambda_1, \lambda_2$ are weighting
factors that can adjust the priorities of the different optimization goals (for tuning of $\lambda_1$, $\lambda_2$, see the example in Section IV).

\subsection{Group Partition}      
\label{Group Partition}
As there may be subgroups in the group, we proceed to infer the subgroups based on
the identified formula
$\phi(\alpha^{\ast})$ where $\alpha^{\ast}$ minimizes $J_{stl}$, 
\begin{definition} The signature $s\textcolor{black}{_q}(\phi(\alpha^{\ast}),k,t)$ is defined as the
	satisfaction signature of the agent $k$ with respect to the ``\emph{inner logic}'' formula $\phi(\alpha^{\ast})$ at time $t$ \textcolor{black}{in the $q$-th observation trajectory}. If the agent $k$ satisfies $\phi(\alpha^{\ast})$ at time $t$ \textcolor{black}{in the $q$-th observation trajectory}, then the signature is set to 1 at that time point; otherwise, it is set to 0.           
\end{definition}                                 

We need to cluster the agents of the group into subgroups based on the satisfaction signature trajectories of different agents.
For a given set of $n$ elements, the number of all possible partitions of the set where each
partition has exactly $n_c$ non-empty subsets is the Stirling's
number of the second kind \cite{Graham1994}. The search over all possible partitions of a set is a NP-complete                                               
problem, and the calculation soon becomes intractable when the number of
elements in the set increases. In order to reduce the calculation, we further look into two kinds of relationships:
complementarity and similarity. \\                       

\textcolor{black}{We come back to the furniture moving scenario and assume that there are two subgroups of people who are \textcolor{black}{moving the furnitures}, $\{1,2,3,4\}$ and $\{5,6,7,8\}$.      \\
	Case 1:\\
	The two subgroups take turns to move the \textcolor{black}{furnitures} from Region 1 to Region 2. For example, if the people in subgroup $\{1,2,3,4\}$ move the furnitures for one hour, then the people in subgroup $\{5,6,7,8\}$ will move the furnitures for the next hour. Therefore, people in the same subgroup behave similarly. \\
	Case 2:\\
	There are people from both the two subgroups who move the furnitures from Region 1 to Region 2. For example, if there are always one person from subgroup $\{1,2,3,4\}$ and two people from subgroup $\{5,6,7,8\}$ who move the furnitures for one hour, then there will be always two people from subgroup $\{1,2,3,4\}$ and one person from subgroup $\{5,6,7,8\}$ who move the furnitures for the next hour. In this case, people in the same subgroup behave complementarily in the sense that a certain number of people in the subgroup should perform the task.}\\

Overall, both complementary and
similar relationships can lead to interesting group behaviors,
but with their different nature they should be dealt with
\textcolor{black}{differently}.                                       

\subsubsection{Group Partition Based on Similarity}
A lot of clustering methods are based on similarity. For example,
k-means clustering is frequently used in partitioning $n$ observations into $k$ clusters, 
where each observation belongs to the cluster with the nearest
mean. However, its
performance can be distorted when clustering high-dimensional data \cite{SWJF2012}. \textcolor{black}{As we cluster different agents based on their satisfaction signatures at different time points (which is high-dimensional when dealing with lengthy time-series data), we need to use some \textcolor{black}{other methods}. One way is to represent the agents in the group as vertices of a weighted hypergraph (a hypergraph is
	an extension of a graph in the sense that each hyperedge can connect more than two vertices) and represent the relationship among different agents as hyperedges. Then the clustering problem is transformed to a hypergraph-partitioning problem where a number of graph-partitioning software packages can be utilized. For example, hMETIS is a software package that can partition large hypergraphs in a fast and efficient way \cite{Karypis1998}. hMETIS can
	partition the vertices of a hypergraph, such that the number of hyperedges connecting
	vertices in different parts is minimized (minimal cut).  \textcolor{black}{The complexity of hMETIS
		for a k-way partitioning is $O((V+E)\log k)$ where V is the number of vertices and E is the number of edges \cite{Karypis1997}.} In this paper, we modify the method in \cite{Karypis1997} which uses frequent item sets found by the association rule algorithm as hyperedges.} Apriori algorithm \textcolor{black}{\cite{TX2013}} is often used in finding association rules in data mining. It proceeds by identifying the frequent individual items\footnote{A frequent individual item is an item that appears sufficiently often through time.} and extending them to larger and larger frequent item sets\footnote{A frequent item set is an item set whose items simultaneously appear sufficiently often through time.}.  In this work, we consider the different agents as items and an item ``appears'' whenever the satisfaction signature of the agent is 1.

\begin{definition} We define the relative support \textcolor{black}{$supp(k,\phi(\alpha^{\ast}))$} of \textcolor{black}{agent} $k$ with respect to the ``\emph{inner logic}'' formula $\phi(\alpha^{\ast})$ as the proportion of satisfaction signatures of the agent $k$ with respect to $\phi(\alpha^{\ast})$ that are not zero, as shown below:
	\begin{equation}
	supp(k,\phi(\alpha^{\ast})) \triangleq \frac{1}{\textcolor{black}{{\displaystyle\sum_{q=1}^{z}l_{\phi(\alpha^{\ast}),q}}}}\textcolor{black}{\displaystyle\sum_{q=1}^{z}}\sum_{j=1}^{l_{\phi(\alpha^{\ast}),q}}s\textcolor{black}{_q}(\phi(\alpha^{\ast}),k,j)
	\end{equation}         
\end{definition}

\begin{table}[h]
	\centering \caption{$s(\phi(\alpha^{\ast}),k,t)$ for 8 agents and 8 time points}
	\label{my-label}
	\begin{tabular}{|c|c|c|c|c|c|c|c|c|c|}
		\hline
		& $t_1$    & $t_2$    & $t_3$    & $t_4$    & $t_5$    & $t_6$ & $t_7$    & $t_8$ &\tabincell{c}{\textcolor{black}{$supp$}\\$\textcolor{black}{(k,\phi(\alpha^{\ast}))}$}
		\\ \hline Agent~$1$ & 1      & 1      & 0      & 0      & 1      & 1  & 0      & 0   & 0.5 
		\\ \hline Agent~$2$ & 1      & 1      & 0      & 0      & 1      & 1  & 0      & 0   & 0.5
		\\ \hline Agent~$3$ & 1      & 1      & 0      & 0      & 1      & 1 & 0      & 0    & 0.5
		\\ \hline Agent~$4$ & 0      & 0      & 1      & 1      & 0      & 0  & 1      & 1  & 0.5
		\\ \hline Agent~$5$ & 0      & 0      & 1      & 1      & 0      & 0  & 1      & 1  & 0.5
		\\ \hline Agent~$6$ & 0      & 0      & 1      & 1      & 0      & 0  & 1      & 1  & 0.5
		\\ \hline Agent~$7$ & 1      & 0      & 0      & 0      & 0      & 0   & 0      & 0  & 0.125
		\\ \hline Agent~$8$ & 0      & 0      & 0      & 1      & 0      & 0   & 0      & 0  & 0.125
		\\ \hline
	\end{tabular}                                              \label{my-label}                   
\end{table}

We give a simple example of 8 agents and \textcolor{black}{1 observation trajectory of} 8 time points (representing 8 consecutive hours) \textcolor{black}{for each agent} in the furniture moving scenario with the signature
$s(\phi(\alpha^{\ast}),k,t)$ listed in Table \ref{my-label}.
We first put all agents that are
identified as frequent individual items in $S_f$,
as shown below:         
\begin{align}
\begin{split}                                
& S_f\triangleq\{k \in S~|~ supp(k,\phi(\alpha^{\ast})) >  {\rm minsup}\} \\  
\end{split} 
\end{align}
where ${\rm minsup}$ is a small positive number as a threshold for 
defining frequent item sets. \textcolor{black}{It can be seen from Table \ref{my-label} that agent 7 and agent 8 do not perform the ``\emph{inner logic}'' task as frequently as the other players, so we set ${\rm minsup}=0.2$ to exclude them from the partitioning process (in similarity relationships all the agents in each subgroup are expected to perform the task frequently)}. \textcolor{black}{In this case, $S_f=\{1, 2, 3, 4, 5, 6\}$}.

\begin{definition} \textcolor{black}{The signature $s\textcolor{black}{_q}(\phi(\alpha^{\ast}),e,t)$ is defined as the
		satisfaction signature of the set $e$ with respect to the ``\emph{inner logic}'' formula $\phi(\alpha^{\ast})$ at time $t$ \textcolor{black}{in the $q$-th observation trajectory}. If all the agents in the set $e$ satisfy $\phi(\alpha^{\ast})$ at time $t$ \textcolor{black}{in the $q$-th observation trajectory}, then the
		signature $s\textcolor{black}{_q}(\phi(\alpha^{\ast}),e,t)$ is set to 1 at that time point; otherwise, it is set to 0}.            
\end{definition}

\begin{definition} \textcolor{black}{We denote the relative support $supp(e,\phi(\alpha^{\ast}))$ of a set $e$ with respect to the ``\emph{inner logic}'' formula $\phi(\alpha^{\ast})$ as the proportion of satisfaction signatures of the set $e$ with respect \textcolor{black}{to} $\phi(\alpha^{\ast})$ that are not zero, as shown below}:
	\begin{equation}
	\textcolor{black}{supp(e,\phi(\alpha^{\ast})) \triangleq \frac{1}{\textcolor{black}{\displaystyle\sum_{q=1}^{z}}l_{\phi(\alpha^{\ast}),q}}\textcolor{black}{\sum_{q=1}^{z}}\sum_{j=1}^{l_{\phi(\alpha^{\ast})},q}s\textcolor{black}{_q}(\phi(\alpha^{\ast}),e,j)}
	\end{equation}         
\end{definition}

\textcolor{black}{If the relative support of a set $e$ satisfies $supp(e,\phi(\alpha^{\ast}))> {\rm minsup}$, then we assign a hyperedge connecting the vertices (agents) of $e$,  and the weight of hyperedge $e$ is defined as relative support $supp(e,\phi(\alpha^{\ast}))$}: 

\begin{equation}          
\textcolor{black}{Weight(e) = supp(e,\phi(\alpha^{\ast})) }
\end{equation}

\textcolor{black}{The fitness function that measures the \textcolor{black}{quality} of a partition (subgroup) $S_d$ is defined as follows}:     
\begin{align}
\begin{split}
\textcolor{black}{fitness(S_d)}&\textcolor{black}{=\frac{\sum_{e \subset S_d}Weight(S_d)}{\sum_{|e \cap S_d|>0}Weight(S_d)}}
\end{split}
\end{align} 
\textcolor{black}{The fitness function measures the ratio of weights of hyperedges that are within the partition and weights of hyperedges involving
	any vertex of this partition. High fitness value suggests that 
	vertices within the partition are more connected to each other than to other vertices}. 

\textcolor{black}{We find the largest number of subgroups partitioned using hMETIS while the fitness function of each subgroup stays above a given threshold value. In the example, the smallest number of subgroups is 2, and the best partition given by hMETIS is: $\{1, 2, 3\}$
	and $\{4, 5, 6\}$. The fitness value of the two subgroups are both 1, which is the highest possible value of the fitness function. If we increase the number of subgroups to 3, then the best partition is $\{1\}$, $\{2, 3\}$
	and $\{4, 5, 6\}$.  The fitness value of the three subgroups are 0, 0.25 and 1. So it is clear that the best number of subgroups should be 2}.                                            

\subsubsection{Group Partition Based on Complementarity}

\textcolor{black}{In a complementarity relationship, the number of agents in a subgroup that perform the task is expected to be \textcolor{black}{as constant as possible}. For example, the proposition ``at least 40 and at most 50 agents from a subgroup of 100 agents should perform the task'' is deemed more precise than ``at least 10 and at most 90 agents from a subgroup of 100 agents should perform the task''.  One good measure of how far a set of numbers are spread out is the variance. If a subgroup $e$ of agents act complementarily, then the variance of the number of agents in subgroup $e$ that perform the task at different time points should be small}.

\textcolor{black}{We still transform the clustering problem to a hypergraph-partitioning problem. The partitioning procedure and the definition of fitness functions are the same as the similarity relationship approach. The only differences are that we assign every possible set $e$ of vertices as a hyperedge and the weight of hyperedge $e$ is defined as follows}:    

\begin{align}
\begin{split}
\textcolor{black}{Weight(e)} &\textcolor{black}{= 1/(Var(e)+\epsilon)} 
\end{split}
\end{align}
\textcolor{black}{where $\epsilon$ is a small positive number such as $10^{-7}$ to avoid singularity in the case of $Var(e)=0$, and the variance $Var(e)$ is defined as}
\begin{align}
\begin{split}  
Var(e)&=\textcolor{black}{\sum_{q=1}^{z}}\sum_{j=1}^{l_{\phi(\alpha^{\ast}),q}}\bigg(\sum_{k=1}^{n} \mathcal{X}_{k,e}s\textcolor{black}{_q}(\phi(\alpha^{\ast}),k,j)-\frac{1}{\textcolor{black}{\displaystyle\sum_{q=1}^{z}l_{\phi(\alpha^{\ast}),q}}}\\&\textcolor{black}{\sum_{q=1}^{z}}\sum_{j=1}^{l_{\phi(\alpha^{\ast}),q}}\sum_{k=1}^{n} \mathcal{X}_{k,e}s\textcolor{black}{_q}(\phi(\alpha^{\ast}),k,j)\bigg)^2/\textcolor{black}{\sum_{q=1}^{z}}l_{\phi(\alpha^{\ast}),q}
\end{split}
\end{align}
\textcolor{black}{where $\mathcal{X}_{k,e}$ is a binary variable that describes whether vertex (agent) $k$ belongs to hyperedge $e$, i.e. $\mathcal{X}_{k,e}$=1 if vertex (agent) $k$ belongs to hyperedge $e$ and
	$\mathcal{X}_{k,e}$=0 otherwise}.    \\

\begin{table}[h]
	\centering \caption{$s(\phi(\alpha^{\ast}),k,t)$ for 8 agents and 8 time points}
	
	\begin{tabular}{|c|c|c|c|c|c|c|c|c|c|} 
		\hline
		& $t_1$    & $t_2$    & $t_3$    & $t_4$    & $t_5$    & $t_6$ & $t_7$    & $t_8$\\ \hline
		Agent 1 & 1      & 0      & 0      & 0      & 0      & 0   & 0      & 0   \\
		\hline    Agent 2 & 0      & 0      & 0      & 1      & 1      & 0   & 0      & 1
		\\ \hline Agent 3 & 0      & 1      & 0      & 0      & 0      & 1   & 1      & 0 
		\\ \hline Agent 4 & 0      & 0      & 1      & 0      & 0      & 0   & 0      & 0  
		\\ \hline Agent 5 & 1      & 0      & 0      & 1      & 1      & 0   & 1      & 0 
		\\ \hline Agent 6 & 0      & 1      & 1      & 1      & 1      & 1   & 1      & 0
		\\ \hline Agent 7 & 1      & 1      & 0      & 0      & 0      & 1   & 0      & 1 
		\\ \hline Agent 8 & 0      & 0      & 1      & 0      & 0      & 0   & 0      & 1
		\\ \hline
	\end{tabular}     
	\label{my-label2}                                                         
\end{table}

\textcolor{black}{In the furniture moving scenario, we give another example of 8 agents and \textcolor{black}{1 observation trajectory of} 8 time points (representing 8 consecutive hours) \textcolor{black}{for each agent} with the signature
	$s(\phi(\alpha^{\ast}),k,t)$ listed in Table \ref{my-label2}. We start from the smallest number of subgroups and the best partition given by hMETIS is: $\{ 1, 2, 3, 4\}$
	and $\{5, 6, 7, 8\}$. The fitness value of the two subgroups are 0.7354 and 0.7364.  If we increase the number of subgroups to 3, then the best partition is $\{7\}$, $\{1, 2,3, 4\}$
	and $\{5, 6, 8\}$. The fitness value of the three subgroups are 0, 0.7354 and 0.5789. So the best number of subgroups is 2.  It can be seen
	from the table that there are always 1 agent from $\{1, 2, 3, 4\}$ and 2 agents from $\{5, 6, 7, 8\}$ that are
	satisfying $\phi(\alpha^{\ast})$ at any time point}.

\subsection{``Outer logic'' CensusSTL \textcolor{black}{Formula Inference}} 

After partitioning the group into several subgroups, we can proceed to
generate the ``\emph{outer logic}'' CensusSTL formula from the census trajectories.

We denote all parameters
that define the ``\emph{outer logic}'' formula $\gamma$ as $\beta$. The inference of the ``\emph{outer logic}'' CensusSTL formula is also a constrained multi-objective optimization problem for finding the best parameters $\beta$ that
describe the formula $\gamma$, \textcolor{black}{and we use Particle Swarm Optimization to find $\beta$.
	In the inference of the ``Outer logic'' formula, we specify the ``\emph{outer logic}'' formula $\gamma$ to be in the form of $\gamma=\Box_{[0,T_{\gamma})}(\gamma_c \Rightarrow \gamma_e)$, with $\gamma_c$                                           
	being the cause and $\gamma_e$ being the effect formula. In this form, we can capture
	causal relationships \textcolor{black}{that are maintained consistently during a time period}.  All the temporal parameters of $\gamma_c$ and $\gamma_e$ chosen in $\mathbb{R}_{\geqslant 0}$ ($T_{\gamma}$ is the length of the time domain of the formula $(\gamma_c \Rightarrow \gamma_e)$ with respect to the census trajectories).}

We consider 8 templates of temporal logic formula $\gamma$:
\subsubsection{\textcolor{black}{Instantaneous Cause Durational Effect}}
\begin{align}                                                   
\begin{split}
\textcolor{black}{\gamma=}
&\textcolor{black}{\Box_{[0,T_c-\tau_2)}(\gamma_{cs} \Rightarrow \Box_{[\tau_1,\tau_2)}\gamma_{es})}
\end{split}
\end{align}
\textcolor{black}{where $T_c$ is the length of the census trajectories, $\gamma_{cs}$ and $\gamma_{es}$ are the cause and effect CensusSTL formulae without temporal operators. This causal relationship means \textquotedblleft\emph{during the next $T_c-\tau_2$ time units, whenever $\gamma_{cs}$ is true, then $\gamma_{es}$ will always be true from the next $\tau_1$ to $\tau_2$ time units}\textquotedblright.}

\subsubsection{\textcolor{black}{Instantaneous Cause Eventual Effect}}
\begin{align}                                                   
\begin{split}
\textcolor{black}{\gamma=}
&\textcolor{black}{\Box_{[0,T_c-\tau_2)}(\gamma_{cs} \Rightarrow \Diamond_{[\tau_1,\tau_2)}\gamma_{es}),}
\end{split}
\end{align}
\textcolor{black}{which means \textquotedblleft\emph{during the next $T_c-\tau_2$ time units, whenever $\gamma_{cs}$ is true, then $\gamma_{es}$ will be true at least once from the next $\tau_1$ to $\tau_2$ time units}\textquotedblright.}

\subsubsection{\textcolor{black}{Instantaneous Cause Eventual Durational Effect}}
\begin{align}                                                   
\begin{split}
\textcolor{black}{\gamma=}
&\textcolor{black}{\Box_{[0,T_c-\tau_2-\tau_3)}(\gamma_{cs} \Rightarrow                    \Diamond_{[\tau_1,\tau_2)}\Box_{[0,\tau_3)}\gamma_{es})}
\end{split}
\end{align}
\textcolor{black}{which means \textquotedblleft\emph{during the next $T_c-\tau_2-\tau_3$ time units, whenever $\gamma_{cs}$ is true, then $\gamma_{es}$ will be true at least once from the next $\tau_1$ to $\tau_2$ time units and maintain to be true for $\tau_3$ time units}\textquotedblright.}

\subsubsection{\textcolor{black}{Instantaneous Cause Persistent Effect}}
\begin{align}                                                   
\begin{split}
\textcolor{black}{\gamma=}
&\textcolor{black}{\Box_{[0,T_c-\tau_2-\tau_3)}(\gamma_{cs} \Rightarrow \Box_{[\tau_1,\tau_2)}\Diamond_{[0,\tau_3)}\gamma_{es})}
\end{split}
\end{align}

\textcolor{black}{which means \textquotedblleft\emph{during the next $T_c-\tau_2-\tau_3$ time units, whenever $\gamma_{cs}$ is true, then from the next $\tau_1$ to $\tau_2$ time units $\gamma_{es}$ will be true at least once every $\tau_3$ time units}\textquotedblright.}

\subsubsection{\textcolor{black}{Durational Cause Durational Effect}}
\begin{align}                                                   
\begin{split}
\textcolor{black}{\gamma=}
&\textcolor{black}{\Box_{[0,T_c-\tau_3)}(\Box_{[0,\tau_1)}\gamma_{cs}\Rightarrow \Box_{[\tau_2,\tau_3)}\gamma_{es})}
\end{split}
\end{align}
\textcolor{black}{which means \textquotedblleft\emph{during the next $T_c-\tau_3$ time units, whenever $\gamma_{cs}$ is true for $\tau_1$ time units, then $\gamma_{es}$ will always be true from $\tau_2$ to $\tau_3$ time units}\textquotedblright.}

\subsubsection{\textcolor{black}{Durational Cause Eventual Effect}}
\begin{align}                                                   
\begin{split}
\textcolor{black}{\gamma=}
&\textcolor{black}{\Box_{[0,T_c-\tau_3)}(\Box_{[0,\tau_1)}\gamma_{cs} \Rightarrow \Diamond_{[\tau_2,\tau_3)}\gamma_{es})}
\end{split}
\end{align}
\textcolor{black}{which means \textquotedblleft\emph{during the next $T_c-\tau_3$ time units, whenever $\gamma_{cs}$ is true for $\tau_1$ time units, then $\gamma_{es}$ will be true at least once from $\tau_2$ to $\tau_3$ time units}\textquotedblright.}

\subsubsection{\textcolor{black}{Durational Cause Eventual Durational Effect}}
\begin{align}                                                   
\begin{split}
\textcolor{black}{\gamma=}
&\textcolor{black}{\Box_{[0,T_c-\tau_3-\tau_4)}(\Box_{[0,\tau_1)}\gamma_{cs} \Rightarrow \Diamond_{[\tau_2,\tau_3)}\Box_{[0,\tau_4)}\gamma_{es})}
\end{split}
\end{align}
\textcolor{black}{which means \textquotedblleft\emph{during the next $T_c-\tau_3-\tau_4$ time units, whenever $\gamma_{cs}$ is true for $\tau_1$ time units, then $\gamma_{es}$ will be true at least once from $\tau_2$ to $\tau_3$ time units and maintain to be true for $\tau_4$ time units}\textquotedblright.}

\subsubsection{\textcolor{black}{Durational Cause Persistent Effect}}
\begin{align}                                                   
\begin{split}
\textcolor{black}{\gamma=}
&\textcolor{black}{\Box_{[0,T_c-\tau_3-\tau_4)}(\Box_{[0,\tau_1)}\gamma_{cs} \Rightarrow \Box_{[\tau_2,\tau_3)}\Diamond_{[0,\tau_4)}\gamma_{es})}
\end{split}
\end{align}
\textcolor{black}{which means \textquotedblleft\emph{during the next $T_c-\tau_3-\tau_4$ time units, whenever $\gamma_{cs}$ is true for $\tau_1$ time units, then from $\tau_2$ to $\tau_3$ time units $\gamma_{es}$ will be true at least once every $\tau_4$ time units}\textquotedblright.}\\

\begin{definition}
	We define $m(\gamma)$ as the total number of time points at which the formula $\gamma$ is true in the training data set \textcolor{black}{(the time points in different census trajectories are counted separately)}.
\end{definition}  

\begin{definition}
	\textcolor{black}{We define $p(\gamma_c \Rightarrow\gamma_e)$ as the accuracy rate of the CensusSTL formula $\gamma=\Box_{[0,T_{\gamma})}(\gamma_c \Rightarrow \gamma_e)$  in the training data set, and its value is calculated as below}:
	\begin{align}
	\begin{split} 
	p(\gamma_c \Rightarrow\gamma_e)=m(\gamma_c \wedge \gamma_e)/m(\gamma_c)
	\end{split}
	\end{align}
\end{definition} 

\textcolor{black}{$p(\gamma_c \Rightarrow\gamma_e)$ is generally a number between 0 and 1, but in the case of $m(\gamma_c)=0$ its value becomes infinity. To avoid this, we specify $p(\gamma_c \Rightarrow\gamma_e)$ to be -1 when $m(\gamma_c)=0$ in the calculations of the objective functions introduced in the following}.

\textcolor{black}{The optimization has three objectives in general: the first objective is to maximize $100p(\gamma_c \Rightarrow\gamma_e)$, which is the percent value of the accuracy rate of the formula in the training data set; the second objective is to maximize $m(\gamma_c)$ so as to maximize the frequency of the formula $\gamma_c$ in the training data set; the last objective is to make the formula $\gamma$ more precise.}\\

\textcolor{black}{Specifically, for similarity based partitioning, we make the number of agents in a subgroup that perform the task as large as possible (ideally the same as the number of agents in the subgroup), so the optimization is formulated as follows:}\\
\[%
\begin{split}
&\textcolor{black}{\mbox{min} -100p(\gamma_c(\beta) \Rightarrow\gamma_e(\beta))-\lambda'_1 m(\gamma_c(\beta))-\lambda'_2(c_{i1}+c_{j2})}\\
&\textcolor{black}{\mbox{subject~to}}     \\  
&\textcolor{black}{\gamma \in \Phi_{\gamma}},\\
&\textcolor{black}{\gamma_{cs}=n(\phi,S_{i})>c_{i1} (i=1,2,\dots,n_s),}\\ &\textcolor{black}{\gamma_{es}=n(\phi,S_{j})>c_{j2} (j=1,2,\dots,n_s),} 
\end{split}
\]                                        
where $\beta$ (including the temporal parameters and $c_{i1}, c_{j2}$) is the optimization variable, $n_s$ is the number of subgroups partitioned based on similarity, $\lambda'_1$, $\lambda'_2$ are the weighting factors (for tuning of $\lambda'_1$, $\lambda'_2$, see the example in Section IV), $\Phi_{\gamma}$ is the set of the eight templates of $\gamma$. For any of the 8 templates, there are $n_s^2$ different CensusSTL formula $\gamma$ as both $\gamma_c$(which contains $\gamma_{cs}$) and $\gamma_e$ (which contains $\gamma_{es}$) can be about any of the $n_s$ subgroups.\\

\textcolor{black}{In the furniture moving scenario from Tab.~\ref{my-label}, one of the best formula obatained is as follows}:

\begin{align}
\begin{split}
&\textcolor{black}{\Box_{[0,4)}(\Box_{[0,2)}n(\phi,S_{s1})>2 \Rightarrow \Box_{[2,4)}(n(\phi,S_{s2})>2)}
\end{split}
\end{align} 
\\
\textcolor{black}{which means \textquotedblleft\emph{For the next 4 hours, whenever the 3 agents from $\{1, 2, 3\}$
		move the furnitures from Region 1 to Region 2 for 2 hours, then the 3 agents from  $\{4, 5, 6\}$ will be moving the furnitures from Region 1 to Region 2 for the next 2 hours.}\textquotedblright}

\textcolor{black}{For complementarity based partitioning, we make the number of agents in a subgroup that perform the task as \textcolor{black}{constant} as possible, the optimization is formulated as follows:}\\
\[%
\begin{split}
&\textcolor{black}{\mbox{min} -100p(\gamma_c(\beta) \Rightarrow\gamma_e(\beta))-\lambda'_1 m(\gamma_c(\beta))+\lambda'_2\displaystyle\sum_{i=1}^{2n_c}(c'_{i2}-c'_{i1})}\\
&\textcolor{black}{\mbox{subject~to}}\\
&\textcolor{black}{\gamma \in \Phi_{\gamma}},\\
&\textcolor{black}{\gamma_{cs}=\bigwedge_{i=1}^{n_c} \bigg(n(\phi,S_{i})>c'_{i1} \wedge n(\phi,S_{i})<c'_{i2} \bigg)},   \\
&\textcolor{black}{\gamma_{es}=\bigwedge_{i=n_c+1}^{2n_c} \bigg(n(\phi,S_{i})>c'_{i1}\wedge n(\phi,S_{i})<c'_{i2} \bigg)},   \\
&\textcolor{black}{c'_{i2}-c'_{i1}>1(i=1,2,\dots,2n_c),}       \end{split}
\]
\textcolor{black}{where $\beta$ (including the temporal parameters and $c'_{i1}, c'_{i2}$) is the optimization variable, $n_c$ is the number of subgroups partitioned based on complementarity, $\lambda'_1$, $\lambda'_2$ are the weighting factors (for tuning of $\lambda'_1$, $\lambda'_2$, see the example in Section IV)}.

\textcolor{black}{In the furniture moving scenario from Tab.~\ref{my-label2}, one of the best formula obtained is as follows:}
\begin{align}
\begin{split}
&\textcolor{black}{\Box_{[0,6)}(n(\phi,S_{c1})>0 \wedge n(\phi,S_{c1})<2 \wedge n(\phi,S_{c2})>1} \\ &\textcolor{black}{\wedge n(\phi,S_{c2})<3 \Rightarrow \Box_{[0,2)}(n(\phi,S_{c1})>0 \wedge n(\phi,S_{c1})}\\ &\textcolor{black}{<2 \wedge n(\phi,S_{c2})>1\wedge n(\phi,S_{c2})<3))}
\end{split}
\end{align} 
\\
\textcolor{black}{which means \textquotedblleft\emph{For the next 6 hours, whenever there are 1 agent from $\{1, 2, 3, 4\}$
		and 2 agents from $\{5, 6, 7, 8\}$ who are moving the furnitures from Region 1 to Region 2, then for the next 2 hours there will still be 1 agent from  $\{1, 2, 3, 4\}$,
		and 2 agents from $\{5, 6, 7, 8\}$ who move the furnitures from Region 1 to Region 2
	}\textquotedblright}.

\subsection{CensusSTL \textcolor{black}{Formula Validation}} 
\textcolor{black}{The obtained CensusSTL formula is validated in a separate validation data set. The accuracy rate of the formula $\gamma_c \Rightarrow \gamma_e$ in the validation data set is as follows:} 
\begin{align}
\begin{split}
& \textcolor{black}{p_v(\gamma_c \Rightarrow \gamma_e) = m_v(\gamma_c \wedge \gamma_e)/m_v(\gamma_c)}
\end{split}
\end{align} 
\textcolor{black}{where $m_v(\gamma)$ is the total number of time points at which the CensusSTL formula $\gamma$ is true in the validation data set.}

\section{Implementation}
\begin{figure}[ht]
	\centering
	\includegraphics[width=9cm]{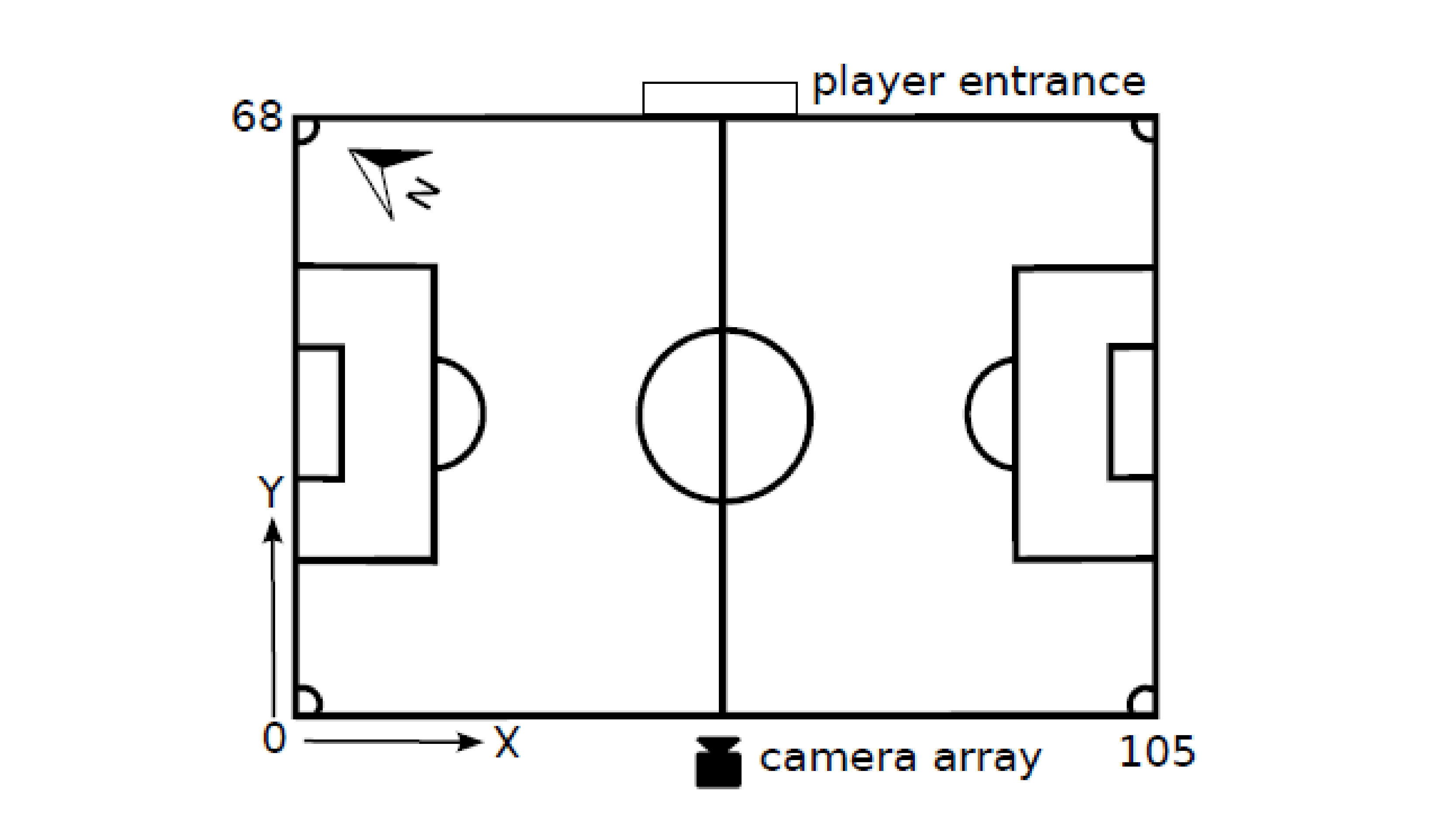}
	\caption{Layout at the Alfheim Stadium \cite{PSJD2014}.}
	\label{fig:soccerfield}%
\end{figure}

In order to test the effectiveness of the algorithm, we consider a \textcolor{black}{dataset from a}
soccer match that happened on November 7th, 2013 \textcolor{black}{between} Troms$\o{}$
IL (Norway) and Anzhi Makhachkala (Russia) at Alfheim stadium in
Troms$\o{}$, Norway. Troms$\o{}$ IL will be referred to as the
home team and the Anzhi Makhachkala as the
visiting team. The players \textcolor{black}{of} the home team are equipped with body-sensors during the whole game. The body-sensor data and video camera data of the
players of the home team are provided in 
\cite{PSJD2014}. The x-axis points southwards parallel \textcolor{black}{with} the long
side of the field, while the y-axis points eastwards parallel with
the short edge of the field, as shown in Fig.~\ref{fig:soccerfield}. The
soccer pitch is $105 \times 68m$ and hence the values for x and y are in
the range of $0 \le x \le 105$ and $0 \le y \le 68$ if the players
are in the field.

\begin{figure}[ht]
	\centering
	\includegraphics[width=9cm]{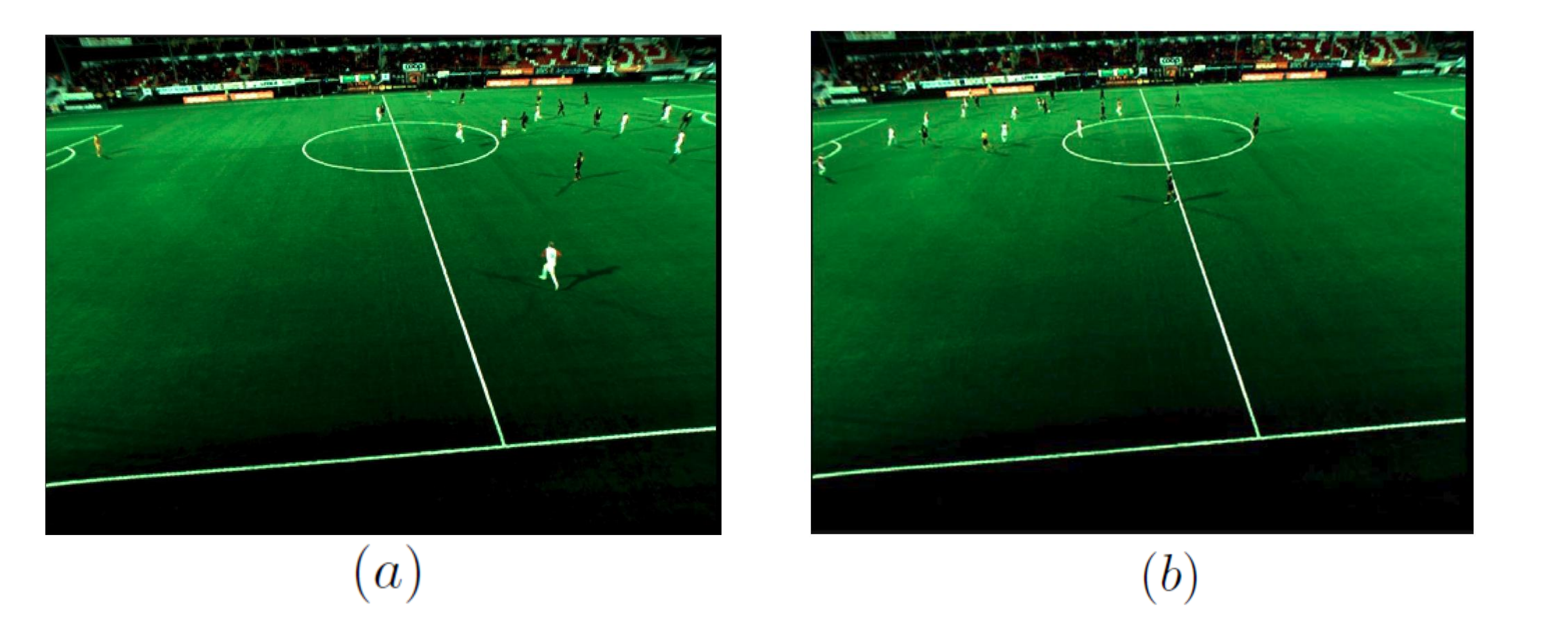}
	\caption{The video capture at (a) 17 minutes 36 seconds and (b) 17 minutes 47 seconds.}
	\label{fig:video}%
\end{figure}

We focus on the situation that a player in the home team is attacking in the visiting team's half field and then the player suddenly runs back to the home field. \textcolor{black}{This usually happens because the ball is intercepted by the visiting team who launches a counterattack and the players in the home team run back for defense. For example, at 17 minutes 36 seconds (as shown in Fig.~\ref{fig:video}(a)) many players are in the
	visiting team's half field, then at 17 minutes 47 seconds most players run back to the home field (as shown in Fig.~\ref{fig:video}(b)).}  We call it a runback situation and we want to derive a CensusSTL formula for the behaviors of different subgroups of the home team. 
\textcolor{black}{As the runback task is a sequential task, we select the following STL formula to be the template for the  ``\emph{inner logic}'' formula:}
\begin{align}
\begin{split}
\textcolor{black}{\phi =} &\textcolor{black}{\Diamond_{[-\tau_3-\tau_4,0]}\phi_t}\\
\textcolor{black}{\phi_t=}&\textcolor{black}{\Box_{[0,\tau_1)}(\textcolor{black}{p_{(red~ region)}}) \wedge \Diamond_{[\tau_2,\tau_3)}\Box_{[0,\tau_4)}(\textcolor{black}{p_{(yellow~ region)}})}
\end{split}                                        
\end{align}
\textcolor{black}{which reads \textquotedblleft\emph{The player stays in the red region for $\tau_1$ seconds, then sometime between $\tau_2$ and $\tau_3$ seconds he arrives in the yellow region and stay there for at least $\tau_4$ seconds}\textquotedblright.}

\textcolor{black}{The a priori regions for the red region ($\mathcal{O}(X_1)$) and yellow region ($\mathcal{O}(X_2)$) are selected symmetrically in the home field and the visiting team's half field near the half-way line, as shown in Fig.~\ref{soccerfield4}(a). As the runback task generally takes less than 12 seconds, we set  $\norm{\phi_t(\alpha)}=\tau_3+\tau_4 \le \tau_{limit}= 12$ in the optimization process.}

Considering that there were substitutions of players in the second half of the match, we focus on the first half of the game.                
\textcolor{black}{The data for the positions of the 10 outfield players (the goalkeeper is excluded) are discontinuous (some data are \textcolor{black}{not available} at certain time intervals). We choose to use the data from the longest time interval with continuous data from 16 minutes and 19 seconds to 29 minutes and 27 seconds (789 seconds in total, sampled at every second) as the training data set for CensusSTL formula inference and use the data from 5 minutes and 13 seconds to 9 minutes and 57 seconds (285 seconds in total) as the validation data set.}
\subsubsection{\textcolor{black}{``\emph{Inner logic}'' Formula Inference}}

\begin{figure*}[ht]
	\centering
	\includegraphics[width=14cm]{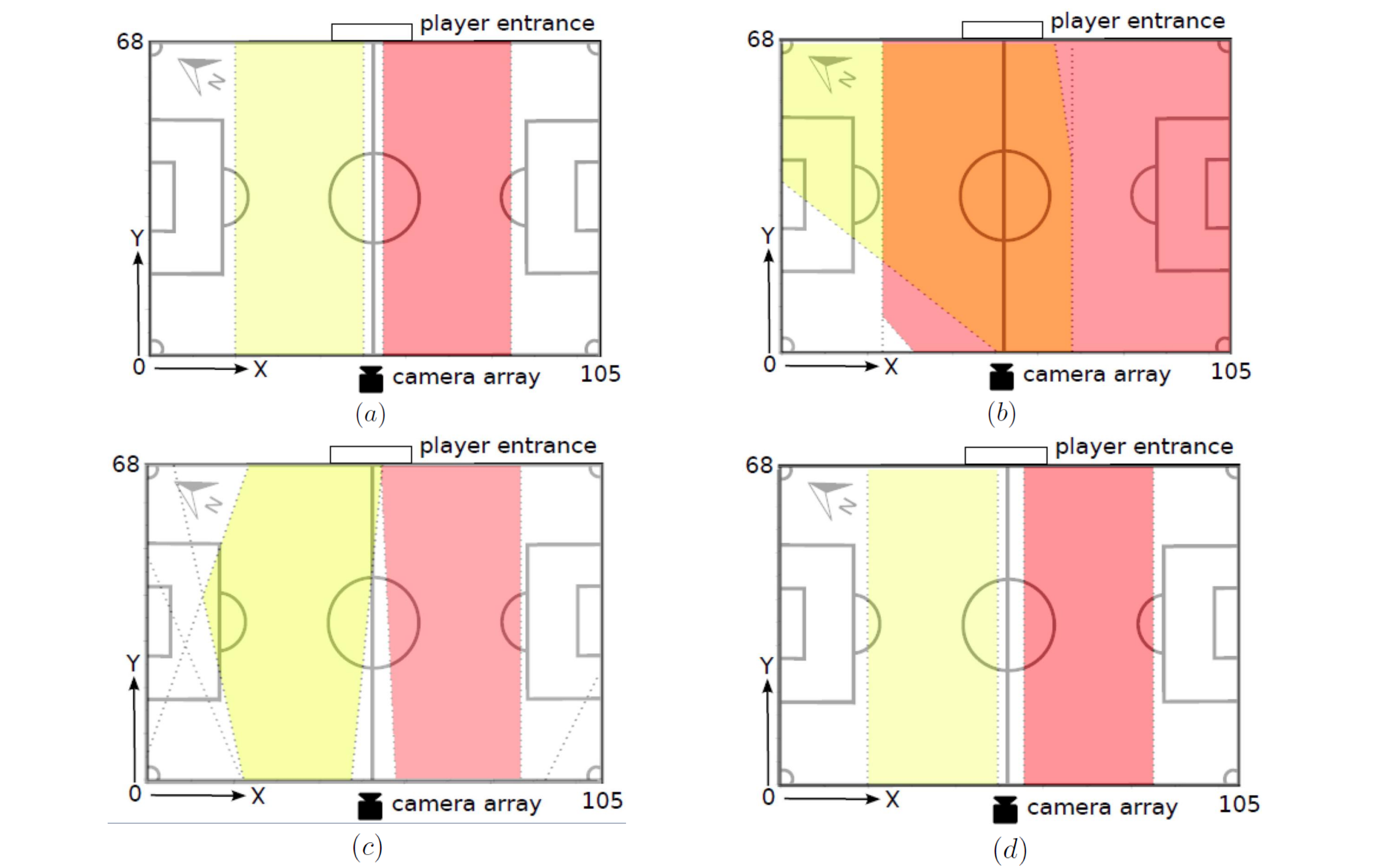}
	\caption{The Alfheim Stadium with (a) a priori regions; (b) obtained regions with $\lambda_2=1$; (c) obtained regions with $\lambda_2=40$; (d) obtained regions with $\lambda_2=100$.}
	\label{soccerfield4}%
\end{figure*}

\textcolor{black}{In cost function (\ref{innercost}), we set $\lambda_1$ in  to be 1, $\lambda_2$ to be 1, 40, 100 and the results are listed in Tab.~\ref{tuning}. When $\lambda_2=1$, there are many times (6788 times) that the ``\emph{inner logic}'' formula is true, but the obtained predicates are very different from the a priori predicates (Hausdorff distances being 986.7829 and 400). Besides, the red region and the yellow region overlap in the middle (as shown in Fig.~\ref{soccerfield4}(b)), which leads to very ambiguous result as the player can just stay in the overlapped region and not actually ``run back''.   When $\lambda_2=100$,  the obtained predicates are almost the same as the a priori predicates (as shown in Fig.~\ref{soccerfield4}(d), Hausdorff distances being 0.3743 and 0.5104), but there are much fewer times (810 times) the ``\emph{inner logic}'' formula is true. In comparison, when $\lambda_2=40$, there are 968 times the ``\emph{inner logic}'' formula is true and the obtained regions remains similar with the a priori regions with no overlap in the middle (as shown in Fig.~\ref{soccerfield4}(c)).}

\begin{table}[h]
	\centering \caption{Results with different $\lambda_2$}
	\label{tuning}
	\begin{tabular}{|c|c|c|c|c|c|c|c|} 
		\hline
		& $\displaystyle\sum_{k=1}^{n} m(\phi(\alpha),k)$    & \tabincell{c}{$d_{\mathrm H}(\mathcal{O}(X_1),$\\$\mathcal{O}(p_1(\alpha)))$}    & \tabincell{c}{$d_{\mathrm H}(\mathcal{O}(X_2),$\\$\mathcal{O}(p_2(\alpha)))$}  \\ \hline
		$\lambda_2=1$ & 6788      & 986.7829      & 400       \\
		\hline    $\lambda_2=40$ & 968      & 5.5881      & 51.4695 
		\\ \hline $\lambda_2=100$ & 810      & 0.3743     &  0.5104      
		\\ \hline
	\end{tabular}                                                               
\end{table}  

The obtained ``\emph{inner logic}'' formula when $\lambda_1=1, \lambda_2=40$ is as follows:

\begin{align}
\begin{split}
&\phi = \Diamond_{[-12,0]}(\Box_{[0,2)}(\textcolor{black}{p_{(red~ region)}(\alpha^{\ast})}) \wedge \Diamond_{[2,10)}\Box_{[0,2)}\\ &(\textcolor{black}{p_{(yellow~ region)}(\alpha^{\ast})}))\\
\\ & \textcolor{black}{\textcolor{black}{p_{(red~ region)}(\alpha^{\ast})}=(0.99873x + 0.050441y  > 57.2938)\wedge }\\ & ~~~~  \textcolor{black}{(0.36068x -0.93269y  < 67.2938)\wedge (0.91245x +}\\ & ~~~~   \textcolor{black}{0.40919 y > 19.7634) \wedge (x < 86.3008)}
\\ & \textcolor{black}{\textcolor{black}{p_{(yellow~ region)}(\alpha^{\ast})}=(0.97221x + 0.23409y  > 21.7704)\wedge} \\ & ~~~~  \textcolor{black}{(0.87666x -0.48111y < 81.3772)\wedge (0.93448x-0.35601y}\\ & ~~~~ \textcolor{black}{> -2.159) \wedge (0.99436x-0.10605y < 47.2862))}
\end{split}
\end{align}

\begin{figure*}[ht]
	\centering
	\includegraphics[width=18cm]{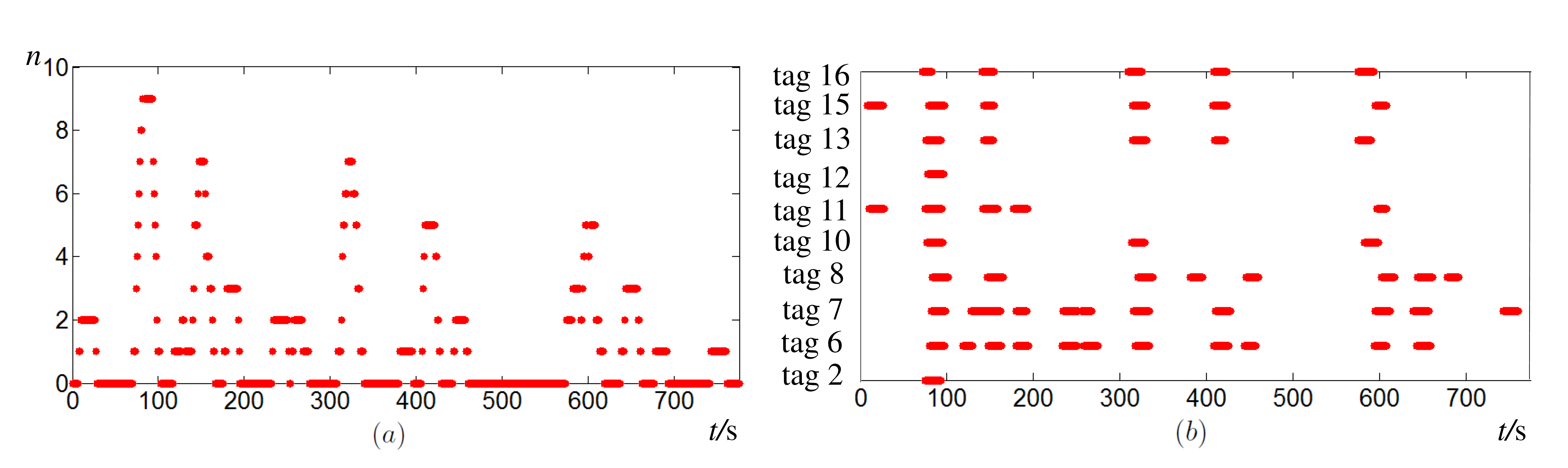}
	\caption{(a) The number of players whose behaviors satisfy $\phi$ in the trainng
		set; (b) The satisfaction signatures of the 10 players in the trainng
		set with respect to $\phi$ (red dots represent that the signature equals one).}
	\label{fig:soccersignal}
\end{figure*}

\begin{table*}[ht]
	\centering \caption{Partition Results based on Similarity}
	
	\begin{tabular}{|c|c|c|c|c|c|c|c|}
		\hline
		& $fitness(S_{s1})$    & $fitness(S_{s2})$    & $fitness(S_{s3})$    & $fitness(S_{s4})$     
		\\ \hline \tabincell{c}{$S_{s1}=\{ tag11, tag13, tag16\}$\\
			$S_{s2}=\{tag6,tag7,tag8,tag15\}$} &  0.3166      &  0.8270     & NA      & NA             
		\\ \hline \tabincell{c}{$S_{s1}=\{tag13, tag16\}$\\
			$S_{s2}=\{tag6,tag7,tag8\}$\\
			$S_{s3}=\{tag11,tag15\}$} & 0.4387      &   0.6943     &  0.2421      & NA           
		\\ \hline \tabincell{c}{$S_{s1}=\{tag6,tag7,tag8\}$\\
			$S_{s2}=\{tag13,tag16\}$\\
			$S_{s3}=\{tag11\}$\\
			$S_{s4}=\{tag15\}$} &  0.6943      & 0.4387      & 0      & 0    
		\\ \hline 
	\end{tabular}     
	\label{parsim}                  
\end{table*}

\begin{table*}[ht]
	\centering \caption{Partition Results based on Complementarity.}
	
	\begin{tabular}{|c|c|c|c|c|c|c|c|}
		\hline
		& $fitness(S_{c1})$    & $fitness(S_{c2})$    & $fitness(S_{c3})$    & $fitness(S_{c4})$   
		\\ \hline \tabincell{c}{$S_{c1}=\{tag6,tag7, tag13,tag15\}$\\
			$S_{c2}=\{tag2,tag8,  tag10,tag11,tag12,tag16\}$} & 0.4256      & 0.6776    & NA      & NA         
		\\ \hline \tabincell{c}{$S_{c1}=\{tag6,tag8,tag11,tag13\} $\\
			$S_{c2}=\{tag2,tag10, tag12, tag16\}$\\
			$S_{c3}=\{tag7,tag15\}$}     &  0.4205      &  0.5241   & 0.1743    & NA            
		\\ \hline 
	\end{tabular}                                \label{parcom}              
\end{table*}                       

\begin{table*}[ht]                            
	\centering \caption{Results with different $\lambda'_2$}
	
	\begin{tabular}{|c|c|c|c|c|c|c|c|}
		\hline
		& $\gamma_1$	& $m(\gamma_c \wedge \gamma_e)$    & $m(\gamma_c)$  & $p(\gamma_c \Rightarrow \gamma_e)$ 
		\\ \hline $\lambda'_2=0.1$ &\tabincell{c}{$\Box_{[0, 780)}(n(\phi,S_{s1})>0 \Rightarrow$ \\  $\Box_{[7,9)}n(\phi,S_{s2})>0)$} & 80      & 99   & 80.81\%  
		\\ \hline $\lambda'_2=1$ &\tabincell{c}{$\Box_{[0,779)}(n(\phi,S_{s1})>0 \Rightarrow$ \\  $\Box_{[8,10)}n(\phi,S_{s2})>1)$} & 65      & 99   & 67.68\%    
		\\ \hline $\lambda'_2=10$ &\tabincell{c}{$\Box_{[0,779)}(n(\phi,S_{s1})>1 \Rightarrow $ \\  $ \Box_{[8,10)}n(\phi,S_{s2})>2)$} & 34      & 65   & 52.31\% 
		\\ \hline  
	\end{tabular}         
	\label{lambda'}                  
\end{table*}

\begin{table*}[ht]
	\centering \caption{THE RESULTS BASED ON SIMILARITY}
	
	\begin{tabular}{|c|c|c|c|c|c|c|c|}
		\hline 
		& $m(\gamma_c \wedge \gamma_e)$    & $m(\gamma_c)$  & $p(\gamma_c \Rightarrow \gamma_e)$  & $m_v(\gamma_c \wedge \gamma_e)$    & $m_v(\gamma_c)$   & $p_v(\gamma_c \Rightarrow \gamma_e)$
		\\ \hline \tabincell{c}{$\gamma_1=\Box_{[0,780)}(n(\phi,S_{s1})>0 \Rightarrow$ \\  $\Box_{[7,9)}n(\phi,S_{s2})>0)$} & 80      & 99   & 80.81\%   & 21     & 42   & 50\%       
		\\ \hline \tabincell{c}{$\gamma_2=\Box_{[0,755)}(n(\phi,S_{s1})>0 \Rightarrow $ \\  $ \Diamond_{[1,34)}n(\phi,S_{s2})>1)$} & 99      & 99   &100\%   & 42      & 42  & 100\%   \\ \hline	
		\tabincell{c}{$\gamma_3=\Box_{[0,738)}(n(\phi,S_{s1})>0 \Rightarrow$ \\ $\Diamond_{[1,50)}\Box_{[0,1)}n(\phi,S_{s2})>1)$}  & 99     & 99  &100\%   & 40      & 42  & 95.24\%  \\ \hline
		\tabincell{c}{$\gamma_4=\Box_{[0,776)}(n(\phi,S_{s1})>0 \Rightarrow$ \\ $\Box_{[2,3)}\Diamond_{[0,10)}n(\phi,S_{s2})>1)$}  & 91      & 99   & 91.92\%   & 22      & 42  &52.38\%  \\ \hline
		\tabincell{c}{$\gamma_5=\Box_{[0,717)}(\Box_{[0,2)}n(\phi,S_{s1})>1 \Rightarrow$ \\ $\Box_{[70,72)}n(\phi,S_{s2})>1)$}  & 74      & 89   & 83.15\%   & 19      & 38  & 50\%  \\ \hline
		\tabincell{c}{$\gamma_6=\Box_{[0,739)}(\Box_{[0,2)}n(\phi,S_{s1})>0 \Rightarrow$ \\ $\Diamond_{[1,50)}n(\phi,S_{s2})>1)$}  & 94      & 94   &100\%   & 40      & 40  & 100\%  \\ \hline
		\tabincell{c}{$\gamma_7=\Box_{[0,766)}(\Box_{[0,2)}n(\phi,S_{s1})>1 \Rightarrow$ \\ $\Diamond_{[1,20)}\Box_{[0,3)}n(\phi,S_{s2})>1)$}  & 89      & 89   & 100\%   & 27     & 38  & 71.05\%  \\ \hline
		\tabincell{c}{$\gamma_8=\Box_{[0, 777)}(\Box_{[0,3)}n(\phi,S_{s1})>0 \Rightarrow$ \\ $\Box_{[5,7)}\Diamond_{[0,5)}n(\phi,S_{s2})>1)$}  & 75      & 89   & 100\%   & 18      & 38  & 47.37\%  \\ \hline
	\end{tabular}   
	\label{outersim}                        
\end{table*} 

\begin{table*}[ht]
	\centering \caption{THE RESULTS BASED ON COMPLEMENTARITY}
	
	\begin{tabular}{|c|c|c|c|c|c|c|c|}
		\hline
		& \tabincell{c}{$m$\\$(\gamma_c \wedge \gamma_e)$}    & $m(\gamma_c)$  & \tabincell{c}{$p$\\$(\gamma_c \Rightarrow \gamma_e)$}   & \tabincell{c}{$m_v$\\$(\gamma_c \wedge \gamma_e)$}    & $m_v(\gamma_c)$   & \tabincell{c}{$p_v$\\$(\gamma_c \Rightarrow \gamma_e)$} 
		\\ \hline \tabincell{c}{$\gamma_1'=\Box_{[0,719)}(n(\phi,S_{c1})>0 \wedge n(\phi,S_{c1})<4$ \\ $\wedge n(\phi,S_{c2})>0 \wedge  n(\phi,S_{c2})<6   \Rightarrow\Box_{[68,70)}$ \\  $(n(\phi,S_{c1})>0 \wedge n(\phi,S_{c1})<4$\\$\wedge n(\phi,S_{c2})>0\wedge n(\phi,S_{c2})<6))$} & 110     & 138   & 79.71\%   & 34      & 42   & 80.95\%  
		\\ \hline \tabincell{c}{$\gamma_2'=\Box_{[0,767)}(n(\phi,S_{c1})>0 \wedge n(\phi,S_{c1})<4$ \\ $\wedge n(\phi,S_{c2})>0 \wedge  n(\phi,S_{c2})<3 \Rightarrow$ \\  $\Diamond_{[1,22)}(n(\phi,S_{c1})>0 \wedge n(\phi,S_{c1})<2$\\$\wedge n(\phi,S_{c2})>0\wedge n(\phi,S_{c2})<2))$} & 128      & 128   & 100\%   & 36      & 36  & 100\%   \\ \hline	
		\tabincell{c}{$\gamma_3'=\Box_{[0,687)}(n(\phi,S_{c1})>0 \wedge n(\phi,S_{c1})<4$ \\ $\wedge n(\phi,S_{c2})>0 \wedge  n(\phi,S_{c2})<4 \Rightarrow\Diamond_{[0,99)}$ \\  $\Box_{[0,3)}(n(\phi,S_{c1})>0 \wedge n(\phi,S_{c1})<2$\\$\wedge n(\phi,S_{c2})>0\wedge n(\phi,S_{c2})<2))$}  & 126      & 132  & 95.45\%   & 36      & 36  & 100\%  \\ \hline
		\tabincell{c}{$\gamma_4'=\Box_{[0,687)}(n(\phi,S_{c1})>0 \wedge n(\phi,S_{c1})<4$ \\ $\wedge n(\phi,S_{c2})>0 \wedge  n(\phi,S_{c2})<4 \Rightarrow$ \\  $\Box_{[0,2)}\Diamond_{[0,100)}(n(\phi,S_{c1})>0 \wedge n(\phi,S_{c1})$\\$<2\wedge n(\phi,S_{c2})>0\wedge n(\phi,S_{c2})<2))$}  & 128      & 132   & 96.97\%   & 32      & 32  & 100\% \\ \hline
		\tabincell{c}{$\gamma_5'=\Box_{[0, 719)}(\Box_{[0,1)}n(\phi,S_{c1})>0 \wedge n(\phi,S_{c1})$ \\ $<4\wedge n(\phi,S_{c2})>0 \wedge  n(\phi,S_{c2})<6   \Rightarrow$ \\  $\Box_{[68,70)}(n(\phi,S_{c1})>0 \wedge n(\phi,S_{c1})$\\$<4\wedge n(\phi,S_{c2})>0\wedge n(\phi,S_{c2})<6))$} & 124     & 124   & 100\%   & 38      & 38   & 100\%  
		\\ \hline
		\tabincell{c}{$\gamma_6'=\Box_{[0, 691)}(\Box_{[0,1)}n(\phi,S_{c1})>0 \wedge n(\phi,S_{c1})$ \\ $<4\wedge n(\phi,S_{c2})>0 \wedge  n(\phi,S_{c2})<6 \Rightarrow$ \\  $\Diamond_{[5,98)}(n(\phi,S_{c1})>0 \wedge n(\phi,S_{c1})<2$\\$\wedge n(\phi,S_{c2})>0\wedge n(\phi,S_{c2})<2))$} & 118      & 124   & 95.16\%   & 34      & 34  & 100\%   \\ \hline
		\tabincell{c}{$\gamma_7'=\Box_{[0,664)}(\Box_{[0,2)}n(\phi,S_{c1})>0 \wedge n(\phi,S_{c1})$ \\ $<4\wedge n(\phi,S_{c2})>0 \wedge  n(\phi,S_{c2})<6 \Rightarrow$ \\  $\Diamond_{[5,124)}\Box_{[0,1)}(n(\phi,S_{c1})>0 \wedge n(\phi,S_{c1})$\\$<2\wedge n(\phi,S_{c2})>0\wedge n(\phi,S_{c2})<2))$}  & 117      & 117   & 100\%   & 30      & 30  & 100\%  \\ \hline
		\tabincell{c}{$\gamma_8'=\Box_{[0,667)}(\Box_{[0,2)}n(\phi,S_{c1})>0 \wedge n(\phi,S_{c1})$ \\ $<4\wedge n(\phi,S_{c2})>0 \wedge  n(\phi,S_{c2})<6 \Rightarrow$ \\  $\Box_{[5,22)}\Diamond_{[0,100)}(n(\phi,S_{c1})>0 \wedge n(\phi,S_{c1})$\\$<2\wedge n(\phi,S_{c2})>0\wedge n(\phi,S_{c2})<2))$} & 114      & 124  &97.14\%   & 34     & 34  & 100\%  \\ \hline
	\end{tabular}       
	\label{outercom}                   
\end{table*} 

\subsubsection{Group \textcolor{black}{Partition}}
In the second step, we partition the group based on the satisfaction signature trajectories of the 10 players with respect to $\phi$.  The number of players whose behaviors satisfy $\phi$ in the team is shown in Fig.~\ref{fig:soccersignal}(a). The satisfaction signatures of the 10 players with respect to $\phi$ 
are shown in Fig.~\ref{fig:soccersignal}(b).We first partition the group based on similarity. \textcolor{black}{It can be seen from Fig.~\ref{fig:soccersignal}(b) that the players of tag2, tag10, tag12 do not perform the runback task as frequently as the other players, so
	we set $minsup$ to be 0.1 to exclude the 3 players. With $S_f=\{tag6,tag7,tag8,tag11, tag13, tag15, tag16\}$, the partition results are shown in Tab.~\ref{parsim}. We set the threshold for the fitness function of a good partition to be 0.2, the largest number of subgroups that can satisfy the criterion is 3.}

We then partition the group based on complementarity and the partition results are shown in Tab.~\ref{parcom}.  \textcolor{black}{With the same threshold for the fitness function as 0.2, the largest number of subgroups that can satisfy the criterion is 2.}

\subsubsection{\textquotedblleft{Outer Logic}\textquotedblright CensusSTL \textcolor{black}{Formula Inference}}

We first identify the CensusSTL formulae for the subgroups
partitioned based on similarity. \textcolor{black}{We find the best formula for each of the 8 different templates. As $n_s=3$, there are $3^2=9$ different formulae for any of the 8 different templates. In this paper, due to space limitations, we only infer the formula with  $\gamma_{cs}=n(\phi,S_{1})>c_{11}$ and $\gamma_{es}=n(\phi,S_{2})>c_{22}$, the other 8 types of formulae can be inferred similarly. We set $\lambda'_1$ in the objective function to be 1, for the value of $\lambda'_2$, we take $\gamma_2$ as an example and obtained different results as listed in Tab.~\ref{lambda'}. When $\lambda'_2=0.1$, we obtain a formula with $80.81\%$ accuracy rate, but the formula is not very precise as $c_{11}$ and $c_{22}$ are relatively small ($c_{11}$=0, $c_{22}$=0) compared to the number of players in the subgroups. When $\lambda'_2=1$, we obtain a formula which is more precise ($c_{11}$=0, $c_{22}$=1), but the accuracy rate of the formula drops to $67.68\%$. When $\lambda'_2=10$, we obtain a formula which is the most precise ($c_{11}$=1, $c_{22}$=2), and the accuracy rate of the formula drops to $52.31\%$. While in general the user can choose a formula with higher accuracy rate or higher \textcolor{black}{precision} based on the preferences of the user, we choose the formulae with higher accuracy rates in such trade-offs in this paper.}

\textcolor{black}{The CensusSTL formulae for the subgroups
	partitioned based on similarity are listed in Tab.~\ref{outersim}. $\gamma_1, \gamma_4, \gamma_5$ do not have $100\%$ accuracy rate in the training data set and get even lower accuracy rates in the validation data set, so they are not the best formulae for this case. $\gamma_7, \gamma_8$ all have $100\%$ accuracy rate in the training data set, but they drop to lower accuracy rates in the validation data set.  In comparison, $\gamma_2, \gamma_3$ and $\gamma_6$ have good performance in both the training data set and the validation data set, so they are the best formulae for similarity relationships.}     

\textcolor{black}{$\gamma_2$ reads \textquotedblleft\emph{Whenever there are at least 1 player from $\{tag13, tag16\}$ who \textcolor{black}{is} running back, then sometime between the next 1 second and the next 34 seconds at least 2 players from $\{tag6, tag7, tag8\}$ will be running back}\textquotedblright.}

\textcolor{black}{$\gamma_3$ reads \textquotedblleft\emph{Whenever there are at least 1 player from $\{tag13, tag16\}$ who \textcolor{black}{is} running back, then sometime between the next 1 second and the next 50 seconds at least 2 players from $\{tag6, tag7, tag8\}$ will be running back for at least 1 second}\textquotedblright.}

\textcolor{black}{$\gamma_6$ reads \textquotedblleft\emph{Whenever there are at least 1 player from $\{tag13, tag16\}$ are running back for 2 seconds, then sometime between the next 1 second and the next 50 seconds at least 2 players from $\{tag6, tag7, tag8\}$ will be running back}\textquotedblright.}

\textcolor{black}{Next we identify the CensusSTL formulae for the subgroups partitioned  
	based on complementarity. The obtained CensusSTL formulae for the 8 different templates are listed in Tab.~\ref{outercom}.  $\gamma_1'$ is the only formula that does not have a $100\%$ accuracy rate in the validation data set and is not accurate enough in the  training data set as well. The rest formulae all have $100\%$ accuracy rates in the validation data set and have over $95\%$ accuracy rates in the training data set, so they are all good formulae for this case. Among them $\gamma_2', \gamma_5', \gamma_7'$ have $100\%$ accuracy rates in both the training data set and the validation data set, so they are the best formulae for complementarity relationships. Although $\gamma_5'$ is not very precise, it still provides useful information in a relatively strong formula structure (bounded always for both cause and effect formula).}

\textcolor{black}{$\gamma_2'$ reads \textquotedblleft\emph{Whenever there are at least 1 and at most 3 players from $\{tag6,tag7,tag13,tag15\}$
		and at least 1 and at most 2 players from $\{tag2,tag8,tag10,tag11,tag12,tag16\}$ who are running back, then sometime between the next 1 second and the next 22 seconds there will still be exactly 1 player from $\{tag6,tag7,tag13,tag15\}$
		and 1 player from $\{tag2,tag8,tag10,tag11,tag12,tag16\}$ who are running back}\textquotedblright.}\\      

\textcolor{black}{$\gamma_5'$ reads \textquotedblleft\emph{Whenever there are at least 1 and at most 3 players from $\{tag6,tag7,tag13,tag15\}$
		and at least 1 and at most 5 players from $\{tag2,tag8,tag10,tag11,tag12,tag16\}$ who are running back for at least 1 second, then from the next 68 seconds to the next 70 seconds there will still be at least 1 and at most 3 players from $\{tag6,tag7,tag13,tag15\}$
		and at least 1 and at most 5 players from $\{tag2,tag8,tag10,tag11,tag12,tag16\}$ who are running back}\textquotedblright.}\\  

\textcolor{black}{$\gamma_7'$ reads \textquotedblleft\emph{Whenever there are at least 1 and at most 3 players from $\{tag6,tag7,tag13,tag15\}$
		and at least 1 and at most 5 players from $\{tag2,tag8,tag10,tag11,tag12,tag16\}$ who are running back for at least 2 seconds, then sometime between the next 5 seconds and the next 124 seconds there will still be exactly 1 player from $\{tag6,tag7,tag13,tag15\}$
		and 1 player from $\{tag2,tag8,tag10,tag11,tag12,tag16\}$ who are running back for at least 1 second}\textquotedblright.}\\         

\textcolor{black}{In conclusion, we obtain some useful CensusSTL formulae in both similarity and complementarity relationship forms. The choice of the best formula depends not only on the performance in the training and validation data set, but also on the user preferences.}     

\textcolor{black}{On a Dell desktop computer with a 3.20 GHz Intel Xeon CPU and
	8 GB RAM, the ``\emph{inner logic}'' formula inference took 881.2 seconds, the
	group partition took 2 seconds and the ``\emph{outer logic}'' formula inference took 72.4 seconds.}

\section{Conclusion}
In this paper, we develop a novel formal framework for analyzing
group behaviors using the newly defined census signal temporal logic. We
used an inference algorithm to identify subgroups and find the census signal temporal logic fomulae for
different subgroups of a multi-agent system. The inference algorithm is composed of three parts: (i) ``\emph{inner logic}'' formula inference, (ii) group partition based on complementarity and similarity, (iii) ``\emph{outer logic}'' CensusSTL formula inference. Using the trajectories generated \textcolor{black}{from the training
	set}, the algorithm can
discover new temporal-spatial properties about the structure of the system. We apply the algorithm in analyzing a soccer match, but similar approach can be used in the recommender systems, biological systems, multi-agent robot systems, monitoring systems, etc.

\bibliographystyle{IEEEtran}
\bibliography{IEEEabrv,zherefacronym}

\end{document}